\documentclass[final]{cvpr}

\usepackage{times}
\usepackage{epsfig}
\usepackage{graphicx}
\usepackage{amsmath}
\usepackage{amssymb}

\usepackage{subcaption}
\usepackage{mathtools}
\usepackage{algorithm}
\usepackage{algorithmic}
\usepackage{multirow}
\usepackage{enumitem}

\newtheorem{definition}{Definition}

\usepackage[pagebackref=true,breaklinks=true,colorlinks,bookmarks=false]{hyperref}

\def\cvprPaperID{7418} 
\def\confYear{CVPR 2021}
\begin{document}

\title{Automatic Correction of Internal Units in Generative Neural Networks}

\author{Ali Tousi$^{1,}$\thanks{Equal contribution},
\and Haedong Jeong$^{1,2,}$\footnotemark[1],
\and Jiyeon Han$^1$,
\and Hwanil Choi$^1$,
\and Jaesik Choi$^{1,3,}$\thanks{Corresponding Author} 
\and $^1$Korea Advanced Institute of Science and Technology (KAIST), South Korea
\and $^2$Ulsan National Institute of Science and Technology (UNIST), South Korea
\and $^3$INEEJI, South Korea\\
{\tt\small \{ali.tousi, haedong.jeong, j.han, hwanil.choi, jaesik.choi\}@kaist.ac.kr}

}

\maketitle

\begin{abstract}
Generative Adversarial Networks (GANs) have shown satisfactory performance in synthetic image generation by devising complex network structure and adversarial training scheme. Even though GANs are able to synthesize realistic images, there exists a number of generated images with defective visual patterns which are known as artifacts. While most of the recent work tries to fix artifact generations by perturbing latent code, few investigate internal units of a generator to fix them. In this work, we devise a method that automatically identifies the internal units generating various types of artifact images. We further propose the sequential correction algorithm which adjusts the generation flow by modifying the detected artifact units to improve the quality of generation while preserving the original outline. Our method outperforms the baseline method in terms of FID-score and shows satisfactory results with human evaluation.
 
\end{abstract}

\section{Introduction}

\begin{figure*}[t!]
    \centering
        \includegraphics[width=\textwidth]{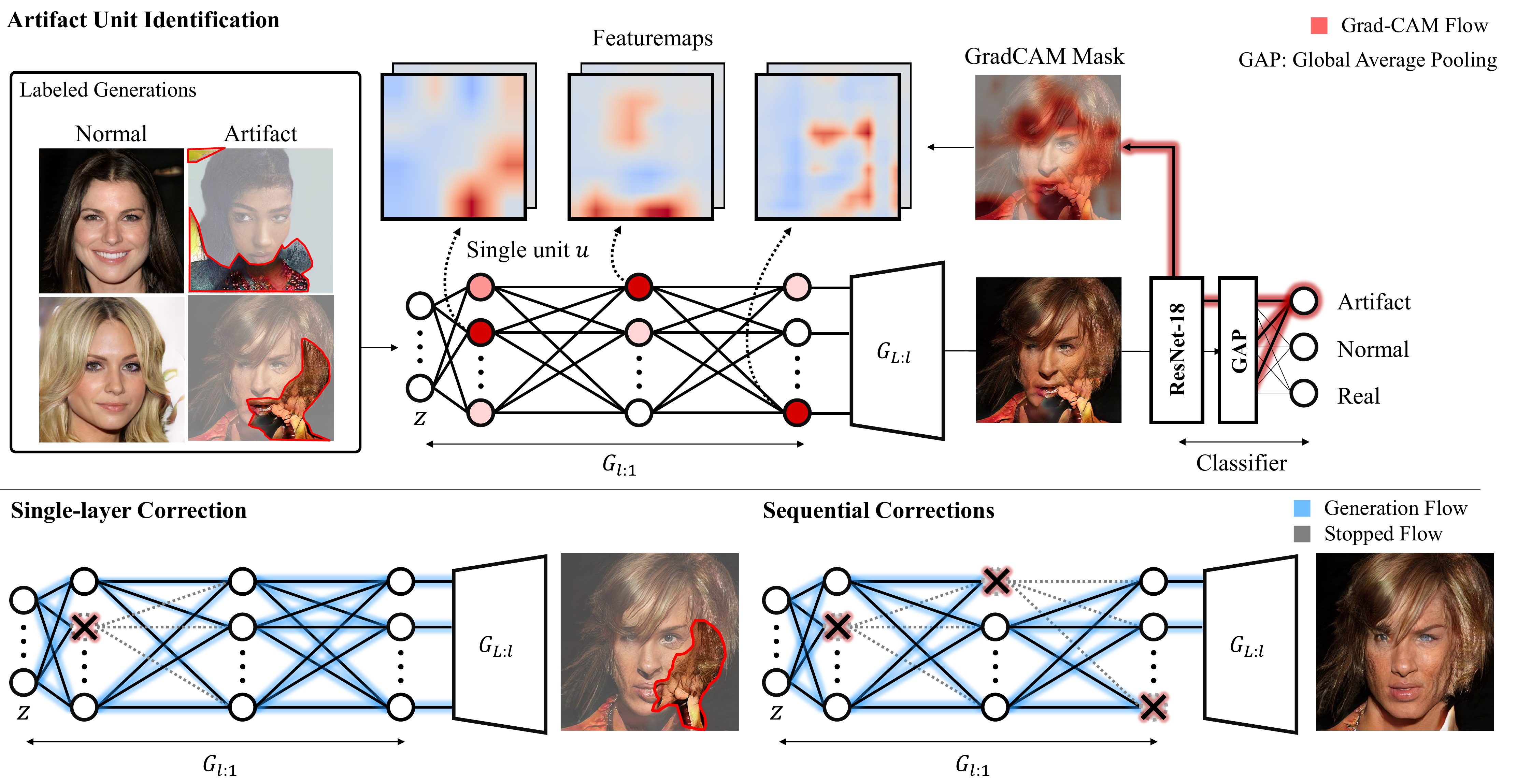}
        \caption{Identification of the artifact units for each layer (top) and the generation flow for two correction methods (bottom). Identification measures the IoU between defective regions (GradCAM mask) and a unit $u$ for artifact generations. The average of IoUs over samples is used as the defective scores in the layer. The sequential correction method adjusts the generation flow of defective units and improves GANs without retraining.}
        \vskip -0.1in
        \label{fig:main}
\end{figure*}

In recent years, GANs have become more powerful in terms of producing photo-realistic images \cite{Karras_2019_CVPR,karras2018progressive} which are often hard to distinguish from real samples. In addition, they have become increasingly better at producing diverse sets of generated samples. These outstanding abilities pave way for GANs to be employed in various real-life domains \cite{pix2pix2017,Mathieu2015Nov}. The main focus of existing work in the GAN domain has been on improving the quality of synthesis by changing the training scheme or devising more complex models. Despite considerable successes, GANs still suffer from producing outputs that contain unrealistic regions, so-called artifacts, which make them unsuitable for being employed in mission-critical applications. As a result, examining the root cause of such phenomena and possible solutions to enhance the overall quality has proved to be important. 

Recent work removes artifact areas by perturbing the latent codes \cite{shen2020interpreting}. Defective units are removed based on understanding the representation of internal units and human annotation \cite{bau2018gan}. Unlike the previous methods, we identify defective units by learning a classifier from annotated samples. Furthermore, by utilizing an obtained explanation map, we devise our own global multi-layer artifact unit ablation scheme which enhances the quality of defective generations while preserving plausible generations.

In our approach, we first annotate randomly sampled generations into two categories, \textit{Normal} and \textit{Artifact}, based on predefined criteria. Then we train a classifier on all the annotated generations and some randomly sampled real images to classify images into their corresponding categories. Our trained classifier can generate an estimated mask for defective regions by employing an explanation method \cite{Selvaraju2016Oct}. By measuring the alignment between an individual internal unit's activation and the defective regions' segmentation mask, we identify units inducing artifacts. In order to correct the artifact areas in the generations, we ablate units with the highest overlap score.

In summary, the contribution of our work is three-fold:
\begin{itemize}
    \item We compile a large dataset of curated flawed generations and provide a comprehensive analysis on artifact generations.
    \item We identify defective units in a generative model by measuring the intersection-over-union (IoU) of the unit's activation map and pseudo artifact region masks obtained by training a simple classifier on our dataset.
    \item We propose an artifact removal method by globally ablating defective units which enhances the quality of artifact samples while maintaining normal samples from drastic change. We further improve the approach by sequentially ablating the defective units throughout consequent layers.
\end{itemize}

\section{Related Work}
\textbf{Generative Adversarial Networks.} Since the introduction of GAN \cite{Goodfellow2014Jun}, the realism and diversity of outputs of the generator have increased steadily \cite{Arjovsky2017Jul,karras2018progressive,Karras_2019_CVPR,Gulrajani2017Mar}. Generally, GAN models take a sampled latent vector and output a synthesized image. While the samples from the original GAN model could be easily identified, recent models produce indistinguishable samples from real data. Despite recent advances, little research has been conducted in order to understand the inner mechanism of a GAN model.

\textbf{Characterizing Deep Networks' Units.} Various techniques have been proposed to examine and understand the internal representation of deep networks \cite{Zeiler2013Nov,Karpathy2015Jun}. Explanatory heatmaps can be used to explain individual network decisions \cite{JMLR:v17:15-618,Selvaraju2016Oct}. The heatmaps visualize which input regions contribute the most to the categorical prediction given by the networks. Recently, \cite{Bau2017Apr} introduces the \textit{Network Dissection} framework for identifying the role of internal units of CNN models by gathering a dataset with semantic concepts and label each hidden unit based on the alignment of its activation map and the concept annotation.

\textbf{Artifacts in Generative Models.} Defective regions can be observed in synthesized images by deep neural networks \cite{Odena2016Oct,Zhang2019Jul,bau2018gan,shen2020interpreting}. \cite{Odena2016Oct} finds the cause of checkerboard artifacts to be in the deconvolution operation and suggests a simple resize-convolution upsampling to resolve this issue. \cite{li2019exposing} exploits the distinctive inherent artifacts that appear in the warping process of generating deep fake videos to be used for detection. \cite{shen2020interpreting} tries to first divide the latent space into normal and artifact regions by training a linear SVM model on manually labeled generations. By moving the latent codes toward the good quality direction obtained from the learned hyperplane, they gradually correct synthesized artifacts images. \cite{bau2018gan} finds artifact-inducing units with/without human supervision. They visualized the highest activation images for each unit and label them as normal or artifact units. Then, they ablate defective units in order to fix the generation. 

\section{Ablation of Artifact Units in GAN}
\begin{figure}[h!]
    \centering
        \includegraphics[width=\columnwidth]{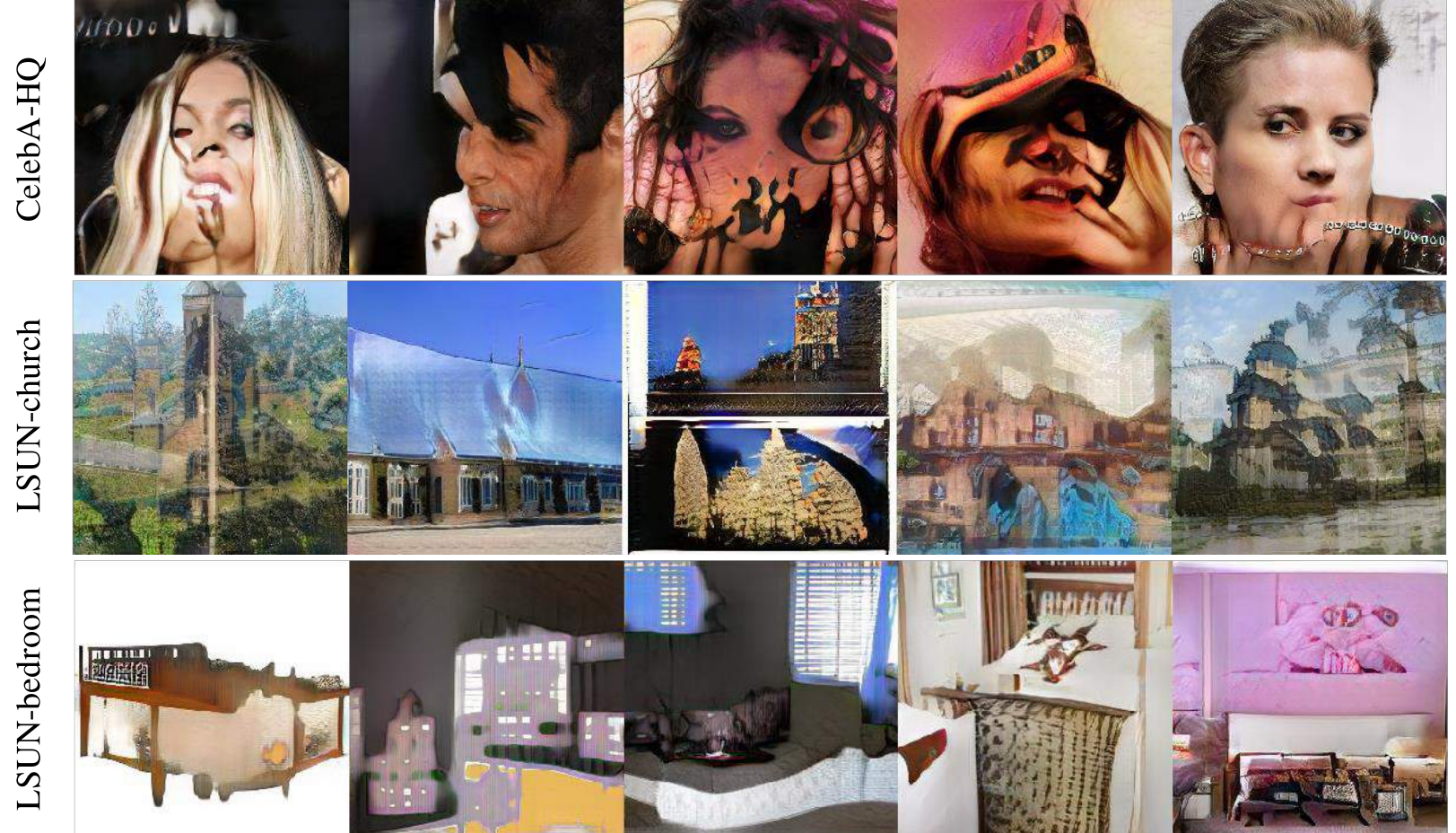}
        \caption{Examples of artifact generations of PGGAN with LSUN-church (top), CelebA-HQ (middle), and LSUN-bedroom (bottom) datasets.}
        \label{fig:artifact_ex}
\end{figure}

The term artifact has been used in the previous work \cite{bau2018gan,Odena2016Oct} to describe the synthesized images which have unnatural (or undesired) visual patterns. Figure \ref{fig:artifact_ex} shows illustrative examples of artifacts, where the church is distorted or some parts of the image are transparent. 
When we label the generations into \textit{artifact} or \textit{normal} generations, we observe one could get non-negligible amount of artifacts as listed in Table \ref{tab:artifact_ratio}.

\begin{table}[ht]
\begin{center}
\begin{tabular}{|c|c|c|c|}
    \hline
\multirow{2}{*}{}&\multirow{2}{*}{\begin{tabular}[c]{@{}c@{}}
    \textbf{CelebA-}\\\textbf{HQ}\end{tabular}} &\multirow{2}{*}{\begin{tabular}[c]{@{}c@{}}
    \textbf{LSUN-}\\\textbf{church}\end{tabular}}&\multirow{2}{*}{\begin{tabular}[c]{@{}c@{}}
    \textbf{LSUN-}\\\textbf{bedroom}\end{tabular}}\\
    &&&\\
    \hline
    \hline
    \begin{tabular}[c]{@{}c@{}}
    Artifact ratio \end{tabular} & 38.71 \% & 16.79 \% & 40.91 \% \\
    \hline
    
\end{tabular}
\end{center}
\caption{The ratio of artifact generations in PGGAN with CelebA-HQ, LSUN-church, and LSUN-bedroom datasets.}
\label{tab:artifact_ratio}
\end{table}

\begin{figure}[h!]
    \centering
    \includegraphics[width=0.48\textwidth]{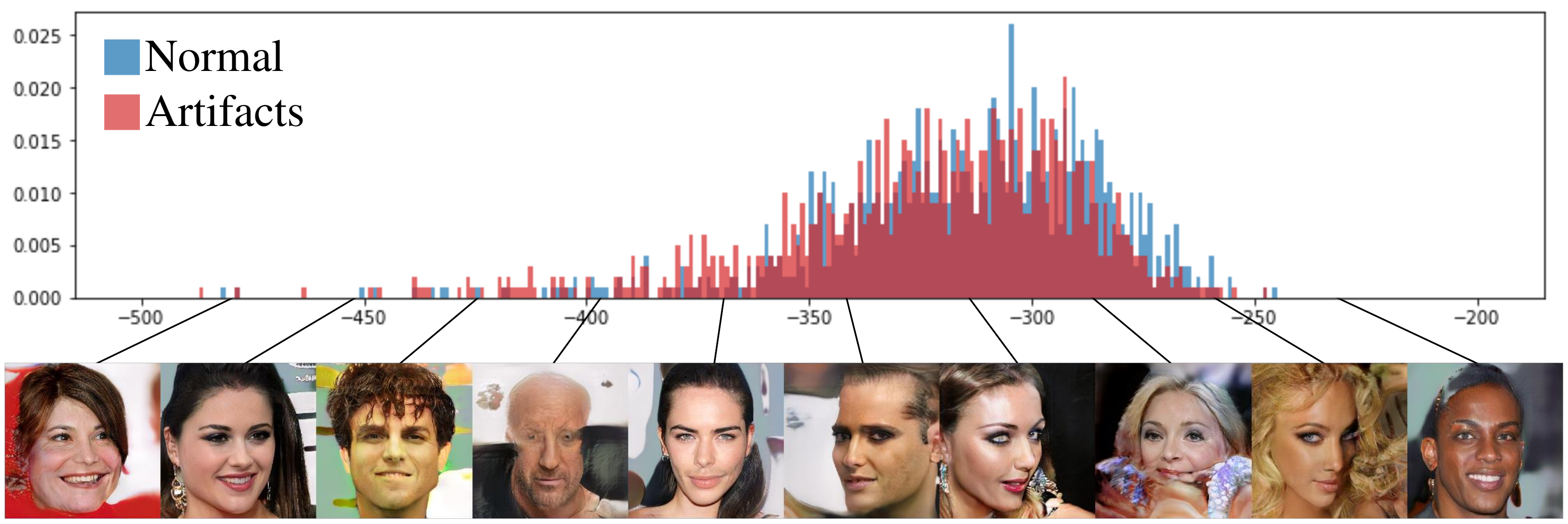}
    \caption{The D-value for each type of input in PGGAN with CelebA-HQ. The black line indicates the corresponding D-value for each generation. The second and the last generations have similar background problems, however, the corresponding D-values are different.}
    \label{fig:dvalue_bad}
\end{figure}

One may suggest using the output of the discriminator, which we call `D-value' for the rest of this paper, as it is a natural metric to distinguish real images and generations. It is common to think the normal generations which look real would have D-values following the distribution of real training images, while artifact generations will be thought to have a distant distribution. However, it was not the case from our observations. 
Figure \ref{fig:dvalue_bad} shows the histogram of D-values for the normal and artifact generations.

It is non-trivial to classify the type of input (normal or artifacts) based on the D-value, because two histograms are overlapped. Also, the level of defectiveness is not proportional to the D-value. These observations show that we need to dig deeper into the network to detect and correct artifacts.

\subsection{FID-Based Artifact Unit Identification}
\begin{figure}[h!]
    \centering
    \includegraphics[width=0.48\textwidth]{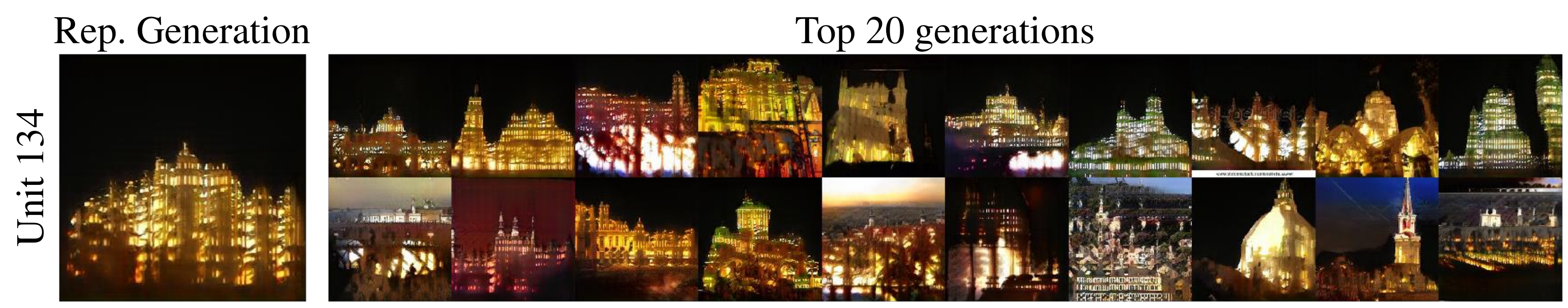}
    \caption{The representative generation and highly activated generations for unit 134 in layer 6 of PGGAN-LSUN church.}
    \label{fig:fid5}
\end{figure}

In the paper \cite{bau2018gan}, artifacts are also studied as one type of object. To get the artifact units in an unsupervised manner, the authors use Fréchet Inception Distance (FID) which is a commonly used metric for measuring performance for generative models \cite{heusel2017gans}. For each featuremap unit, they compute the FID score on 10K images which have the highest activation of the given featuremap unit among 200K generated images. Featuremap units which are highly activated for high-FID image sets are considered artifact units. The authors show that the FID score improves when the top 20 artifact units are ablated. While this automatic correction method has shown overall improvement, our observation shows there is still a possibility to improve the identification of artifact units. Figure \ref{fig:fid5} shows the unit with the 5th highest FID score among 512 units.

Although unit 134 has a high FID score, the related generation concepts seem to be natural. This example illustrates the reason why we need a more elaborate approach for the identification of artifact units. 

\subsection{Classifier-Based Artifact Unit Identification}
In order to identify the internal units which cause high-level semantic artifacts, we hand-label 2k generated images to normal or artifact generations. Then, we build a model to classify our dataset into three categories, namely: artifact, normal, and randomly selected real samples. We employ an image classifier (e.g. ResNet-18) as our feature extraction module and introduce one fully connected layer on top of it for classification. During the training, we keep the parameters of the feature extraction module fixed and only optimize the classifier weights. In order to obtain a finer-level annotation compared to the image label, we apply GradCAM \cite{Selvaraju2016Oct} to provide us with a mask for the defective regions. Such a mask highlights the regions that are effective for the model's decision. This operation is more efficient than hand labeling defective regions as the latter takes a lot of time. 

\begin{figure}[h!]
    \centering
    \includegraphics[width=0.47\textwidth]{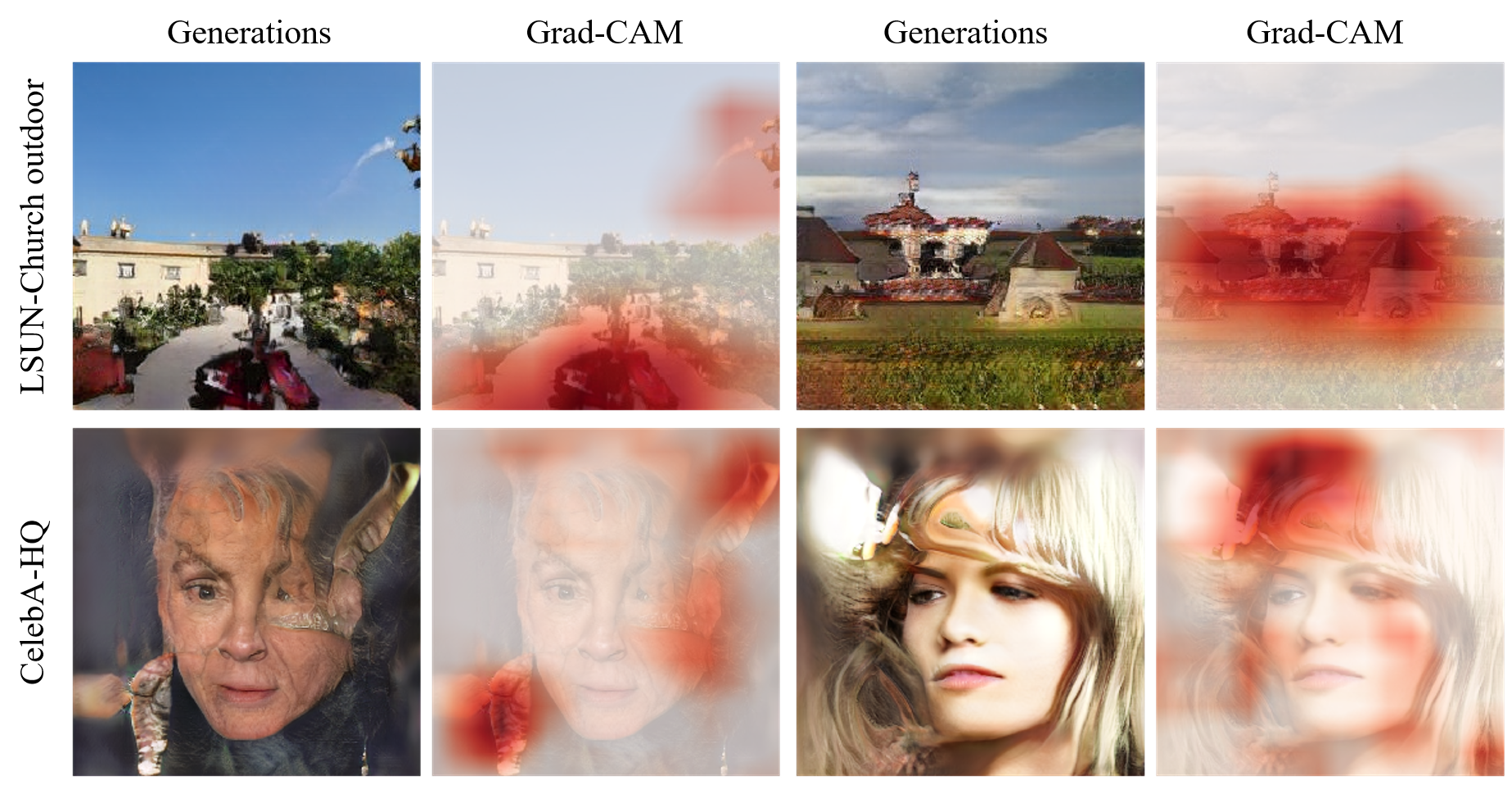}
    \caption{The explanation masks from the GradCAM with artifact class score. The masks focus on the defective regions which the normal generations are hard to contain.}
    \label{fig:gradcam}
\end{figure}

Now that we have a pseudo segmentation ground truth mask for artifact areas, comparing feature activation maps with the mask can reveal the artifact-inducing units.
For this purpose, we follow the principles in the prior work \cite{Bau2017Apr,Fong2018Jan}. For every unit $u$ in the $l$-th layer of the generator, we compute $A_{u}(z_x)\in \mathbb{R}^{H_{l}\times W_{l}}$ which is the activation for a given generation $x$ and the corresponding latent code $z_x$.
We threshold the activation with the quantile $T_u$ such that $P(A_u(z_x) > T_u)  = \tau$\footnote{We empirically set $\tau$ = 0.005 in the experiments.}. This is computed with regard to the feature map distribution of each unit for all images. After applying the threshold, we bilinearly upsample the result to the size of the pseudo artifact segmentation mask $L_{a}(x)$ and compute IoU with $A_u(z_x)$. To make the identification process global, we define the defective score which averages each IoU over artifact generations.

\begin{definition}[\textbf{Defective Score (DS)}]
\label{def:ds}
Let the set of artifact generations $X_a$ and artifact segmentation mask $L_a(x)$ and activation $A_u(z_x)$ of unit $u$ in the $l$-th layer for $x\in X_a$. The defective score of unit $u$ is defined as, 
$$DS_{l,u,a} = \frac{1}{\vert X_a \vert}\sum\limits_{x \in X_a}\frac{\vert A_u(z_x) \cap L_a(x)\vert}{\vert A_u(z_x) \cup L_a(x)\vert}.$$
\end{definition}

The defective scores for all units in the $l$-th layer are defined as $DS_{l,a}=\{ DS_{l,1,a},DS_{l,2,a},...DS_{l,D_l,a} \}$ where $D_l$ is the number of the units in the $l$-th layer.
Finally, we can sort the scores and choose units with higher scores as candidates for ablation.

\begin{figure}[h!]
    \centering
    \includegraphics[width=0.49\textwidth]{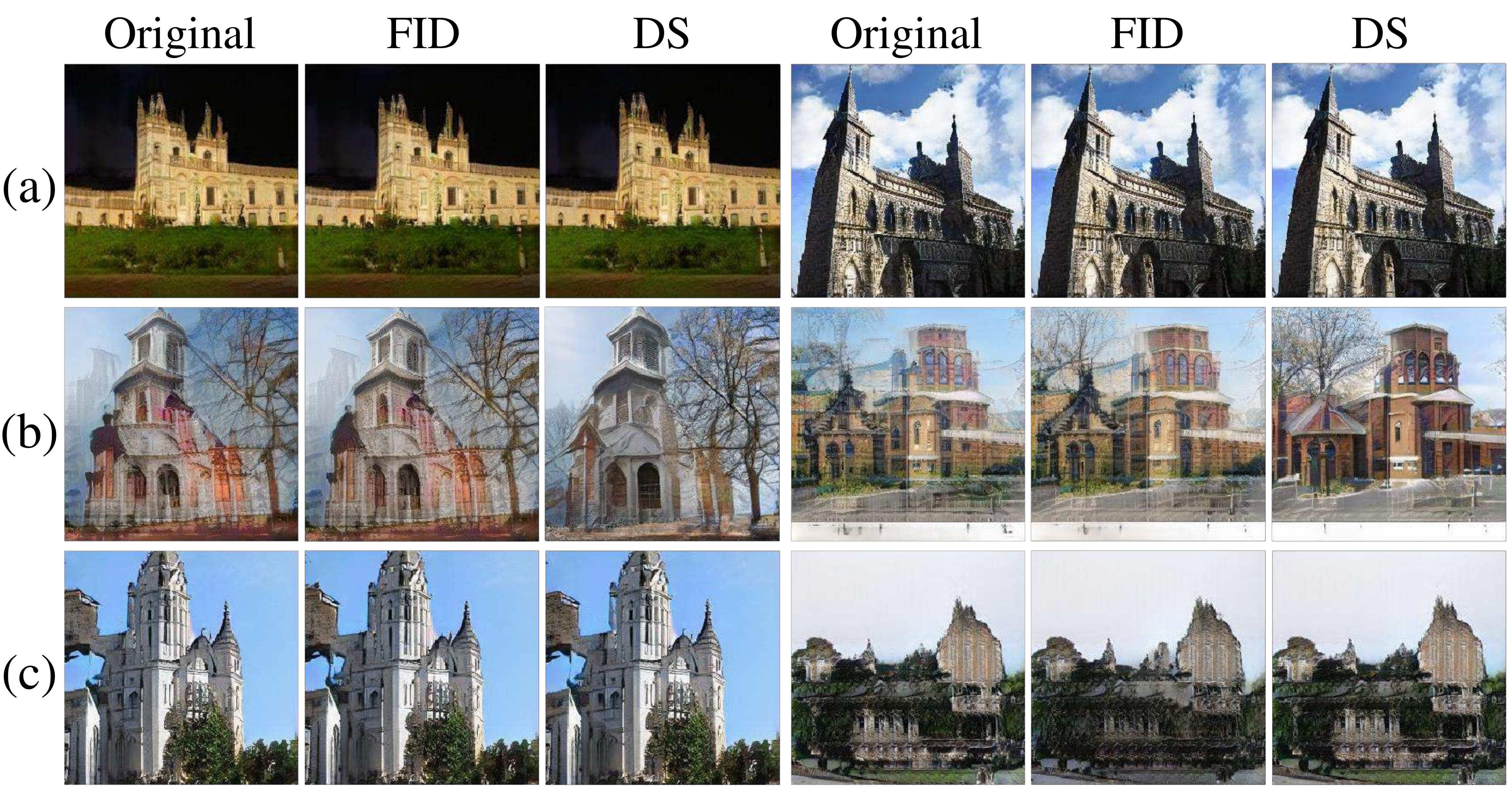}
    \caption{The single-layer ablation results (top 20) for the baseline method (FID) \cite{bau2018gan} and our method (DS). (a) Correction results on normal generations. The ablations from both methods do not affect plausible regions. (b) Our method can remove the shadow effect while maintaining the original outline. (c) Both methods fail to correct the hole on the wall (left) or the texture error (right).}
    \label{fig:basic_correction}
\end{figure}

Figure \ref{fig:basic_correction} denotes the results of ablation with the top 20 units in layer 6 comparing with the FID-based ablation. In the first row, we can identify that both methods barely harm the plausible regions in the generation. It implies that the selected units are related to the defective area and less correlated with the normal generations. In the second row with the shadow artifact, DS-based correction shows more reasonable performance than FID-based correction. However, both methods fail to correct in some cases as in the last row: (1) the hole on the church and (2) the texture error case.

\subsection{Generation Concepts of Unit in GANs}\label{sec:generation_concepts}
\begin{figure}[h!]
    \centering
      \begin{subfigure}[b]{0.48\textwidth}
        \includegraphics[width=\textwidth]{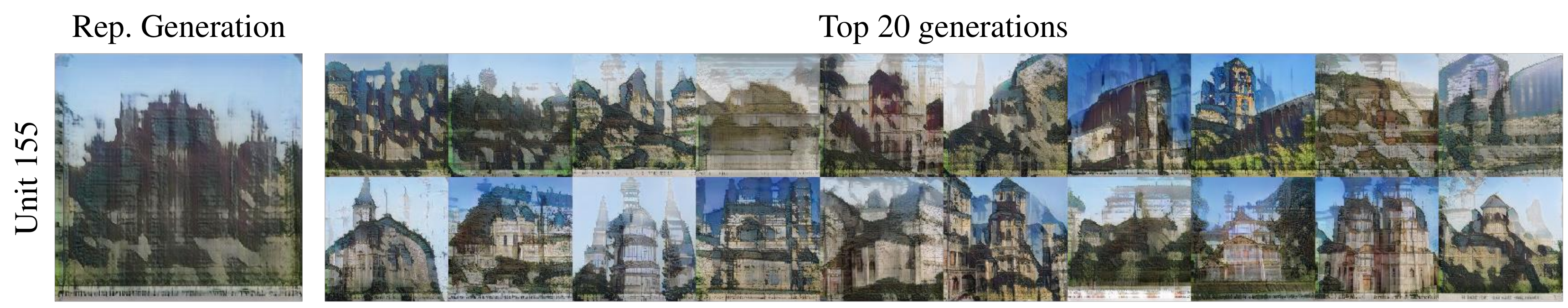}
        \caption{Unit for concrete generation concepts}
        \label{fig:concrete}
     \end{subfigure}
   \begin{subfigure}[b]{0.48\textwidth}
        \includegraphics[width=\textwidth]{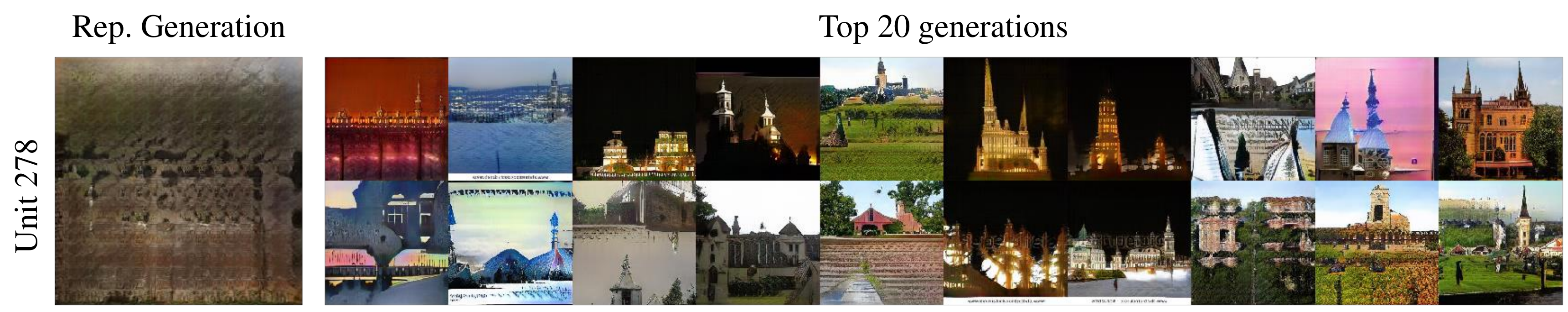}
        \caption{Unit for mixed generation concepts}
        \label{fig:mixed}
     \end{subfigure}
     \caption{The examples of the representative image and highly activated generations for the given unit in layer 6 of PGGAN on LSUN-church. (a) A unit related to concrete generation concepts. (b) A unit that does not have concrete generation concepts. The representative image in this case seems to be blurred.}
     \label{fig:gen_concept}
\end{figure}

From the previous observations, we can identify that a simple unit ablation cannot correct all types of artifacts. To analyze and explain this phenomenon, we investigate and reveal the generation concept of each featuremap unit.
We generate 20k images and select the top 20 images which maximize the magnitude of activation for each unit $u$. Some featuremap units seem to have concrete concepts because the selected images share concrete semantic information (see more examples in Appendix 5). However, for some featuremap units, it is hard to define one clear generation concept. To better identify such concepts, we compute the mean featuremap amongst highly activated images to generate the representative image for each featuremap unit. Figure \ref{fig:gen_concept} denotes the examples of units with concrete/mixed generation concepts.

\begin{figure}[h!]
    \centering
    \includegraphics[width=0.49\textwidth]{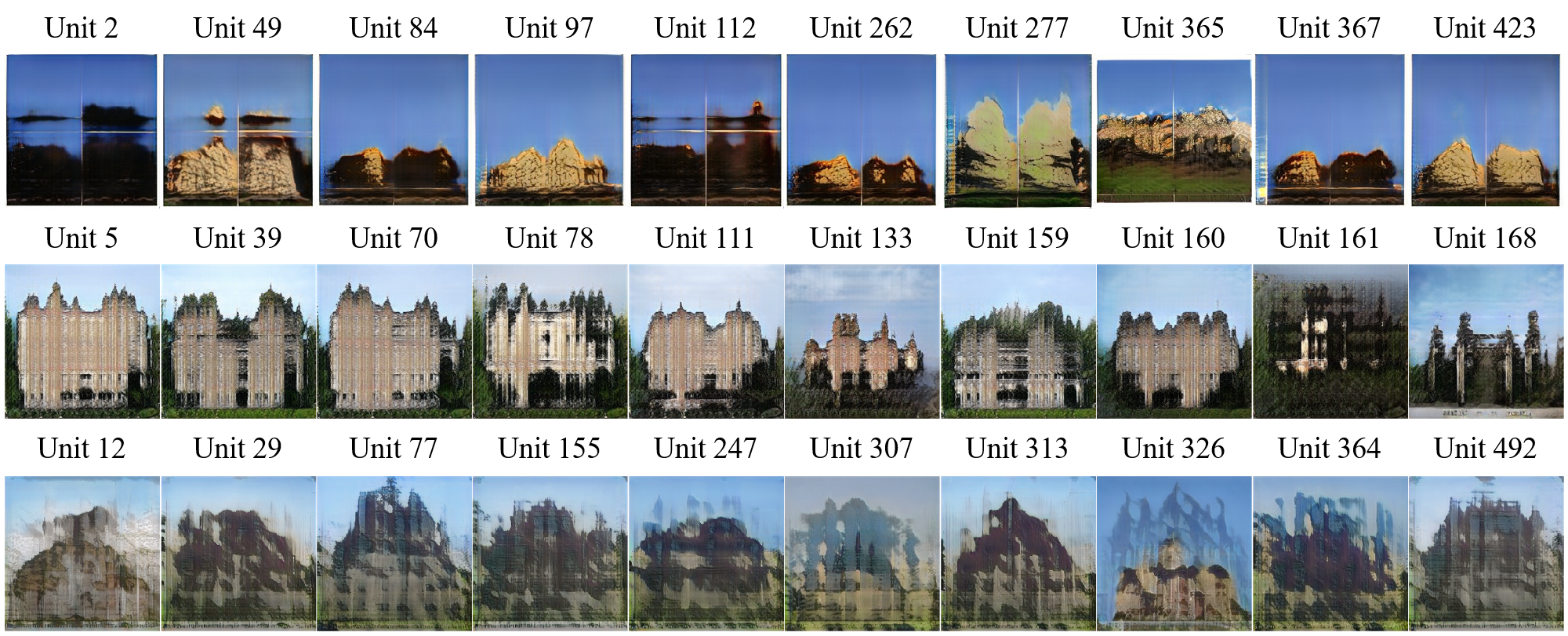}
    \caption{The clusters of the representative generations in layer 6 of PGGAN with LSUN-church. There are multiple units which share the same artifact concept.}
    \label{fig:multiunits}
\end{figure}

\begin{figure}[h!]
    \centering
    \includegraphics[width=0.49\textwidth]{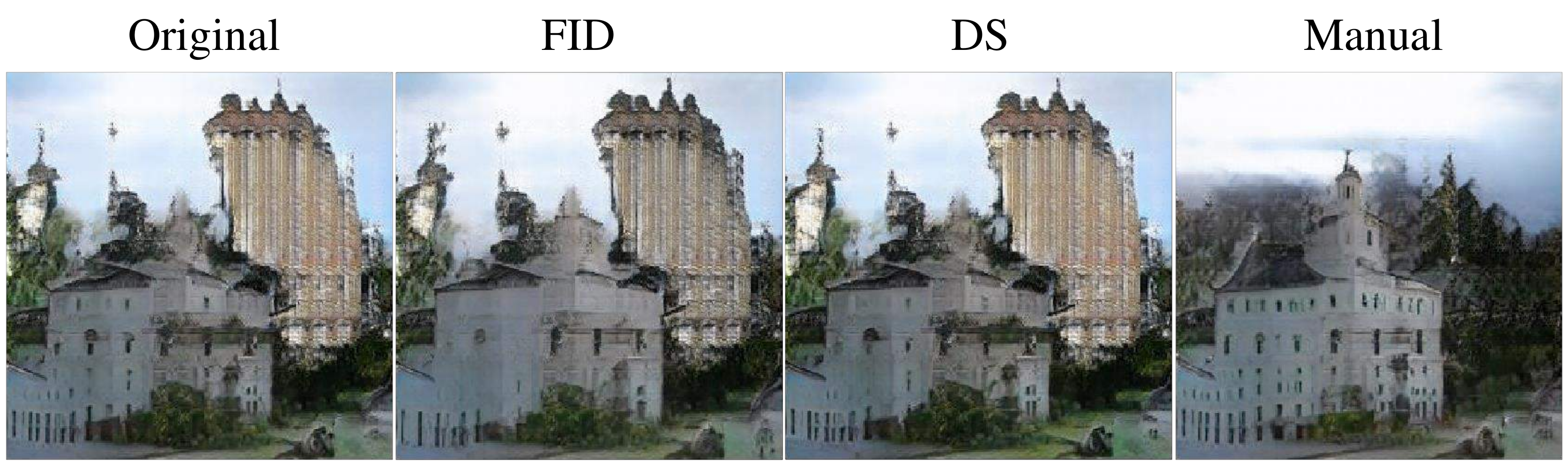}
    \caption{An example of correction results for texture error. In the Manual case, we select and ablate the 15 units related to the texture error.}
    \label{fig:manual_ablation}
\end{figure}

Figure \ref{fig:multiunits} denotes the examples which are clustered by the type of artifacts with the corresponding unit indices. We can clearly identify that each artifact concept is related to multiple units in a hidden layer. It suggests that we need to ablate a set of units which includes most units sharing the defective concepts to ensure the correction performance. Figure \ref{fig:manual_ablation} denotes the correction results comparing with manual unit selection. Because the top 20 units did not include all the texture error units in FID and DS-based correction, they cannot remove the texture error in the generation. However, when we manually ablate 15 units related to the texture error, the correction can be performed efficiently.

\begin{figure}[h!]
    \centering
    \includegraphics[width=0.49\textwidth]{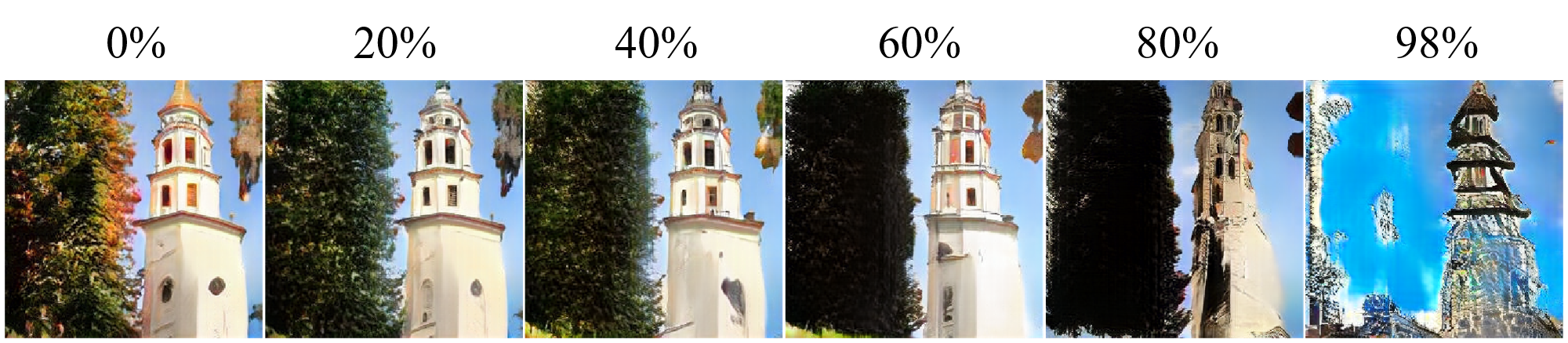}
    \caption{The change in generation over the number of ablation units. Although the size of the stain (top-right) is reduced when increasing the number of ablation units, the quality of trees and the church is degraded at the same time.}
    \label{fig:side_effect}
\end{figure}

\begin{figure}[h!]
    \centering
    \includegraphics[width=0.49\textwidth]{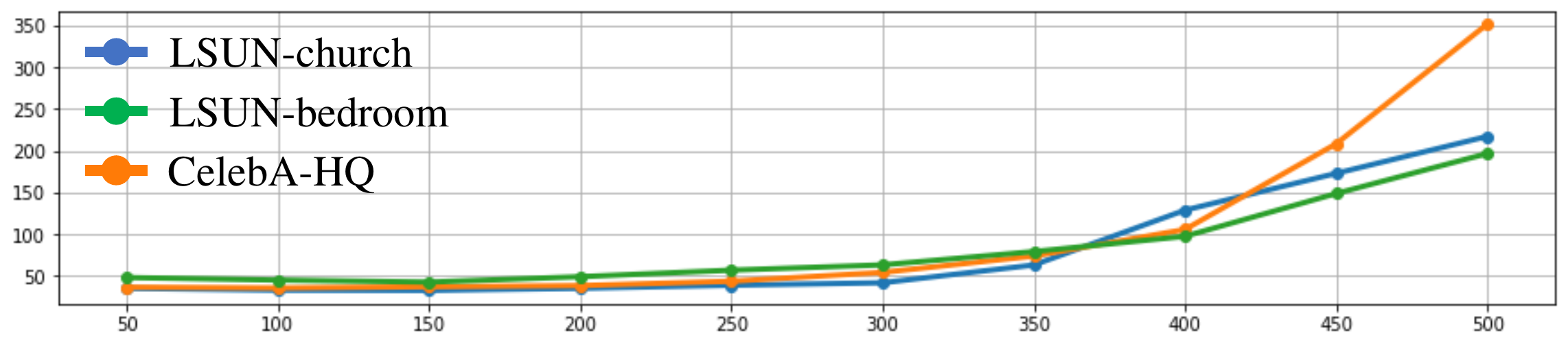}
    \caption{The FID scores on various numbers of ablated units on layer 6 of PGGAN with a various dataset.}
    \label{fig:side_effect_plot}
\end{figure}

Although we can cover various types of artifacts by simply increasing the number of units for ablation, there exists a trade-off between removing artifacts and preserving the normal information in generations. Figure \ref{fig:side_effect} illustrates such a trade-off. When we increase the number of units, it gradually removes the defective spot on the top-right corner in the generation, but heavy degradation also appears (e.g. the quality of trees and the building). In addition, we plot the FID scores according to the number of units ablated at layer 6 for the PGGAN model across various datasets in Figure \ref{fig:side_effect_plot}. We can identify that the FID score increases exponentially when we increase the number of units for ablation.

\section{Sequential Correction of Artifacts}
We propose the sequential correction to improve the quality of generation performance by suppressing the side-effect of one layer ablations. Given generator with $L$ layers, the function of generator $G$ is decomposed into $G(z)=f_{L}(f_{L-1}(\cdots (f_{1}(z)))) = f_{L:1}(z)$, where $z$ is a vector in the latent space $\mathcal{Z}\subset \mathbb{R}^{D_z}$. $h_{l,u}=f_{l:1,u}(\cdot)$ denotes the values of the $u$-th unit in the $l$-th layer with $f_{l:1}(z) \in \mathbb{R}^{D_l\times H_l \times W_l}$. In general, the operation $f_l(\cdot)$ includes linear transformations and a non-linear activation function. For the given query $z$, we sequentially adjust the activation of units in the consequent layers. The detailed procedure is stated in Algorithm \ref{alg:seqcorr}.

From the previous research \cite{jeon2020efficient} that shallow layers handle the abstract generation concepts and deeper layers handle localized information in GANs, we ablate the shallow layers from the first layer to the stopping layer $l<L$. To prevent the loss of semantic characteristics of a generation as pointed in Section \ref{sec:generation_concepts}, we adjust the magnitude of the original featuremaps instead of the simple zero ablation.
Line 5 of Algorithm \ref{alg:seqcorr} states this soft ablation as,

\begin{equation*}
h_{k+1,j} = \lambda (1-DS_{k+1, j, a}) h_{k+1,j}
\end{equation*}
where $\lambda\in[0,1]$ is the scaling factor and $DS_{k+1, j, a}\in[0,1]$ is normalized.
$\lambda (1-DS_{k+1, j, a})$ controls the relative generation flow of selected featuremap units.
Note that if the scaling factor $\lambda=0$, the algorithm performs simple zero ablations in consequent layers.

\begin{algorithm}[h]
\caption{Sequential Correction}\label{alg:seqcorr}
Input: $z_0$: a query, $G(.)=f_{L:1}(.)$: a generator, \\$l$: a stopping layer, $DS_{l:1,a}$: normalized defective scores for each layer, $\lambda$: a scaling factor, $n$: the number of ablated units\\
Output: $X$: the corrected generation

\begin{algorithmic}[1]
\STATE $h_0=z_0$
\FOR{$k \gets 0$ to $l$}
  \STATE $h_{k+1}=f_{k+1:k}(h_k)$
  \FOR{$j \gets$ Top $1$ to Top $n$}
    \STATE $h_{k+1,j}= \lambda(1-DS_{k+1,j,a})h_{k+1,j}$
  \ENDFOR
\ENDFOR
\STATE $X=f_{L:l+1}(h_{l+1})$
\STATE \textbf{return} $X$
\end{algorithmic}
\end{algorithm}

\subsection{Analysis of Sequential Correction}
\begin{figure}[h!]
    \centering
     \begin{subfigure}[b]{0.49\columnwidth}
         \centering
         \includegraphics[width=\textwidth]{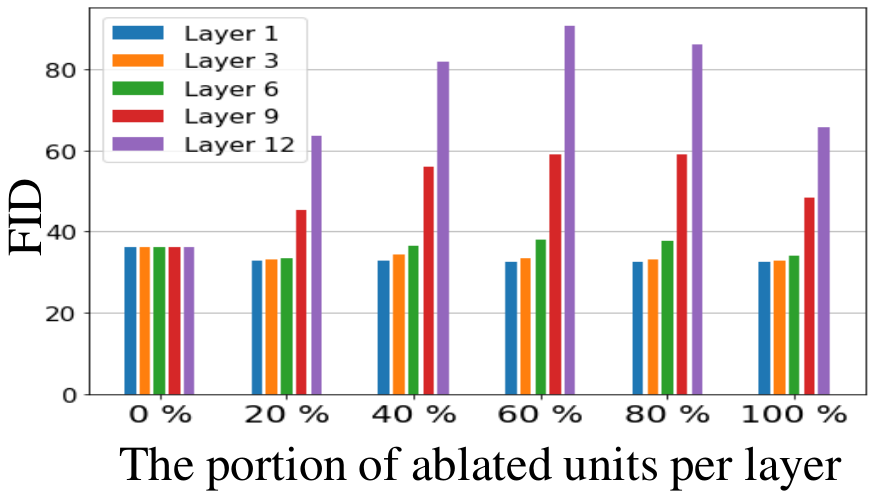}
         \caption{FID for various condition}
    \label{fig:rebuttal_fid}
     \end{subfigure}
     \hfill
    \begin{subfigure}[b]{0.49\columnwidth}
         \centering
         \includegraphics[width=\textwidth]{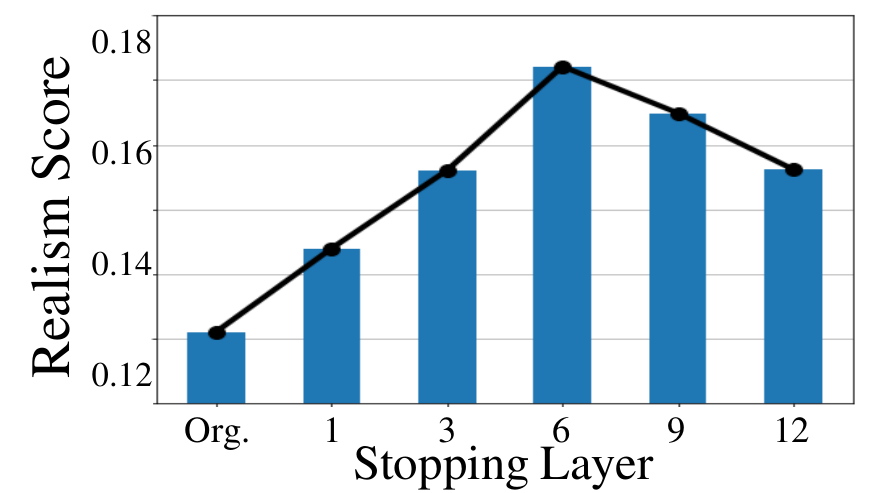}
         \caption{Realism score with n=20\%}
     \label{fig:rebuttal_realism}
     \end{subfigure}
     \hfill
    \begin{subfigure}[b]{0.47\textwidth}
         \centering
         \includegraphics[width=\textwidth]{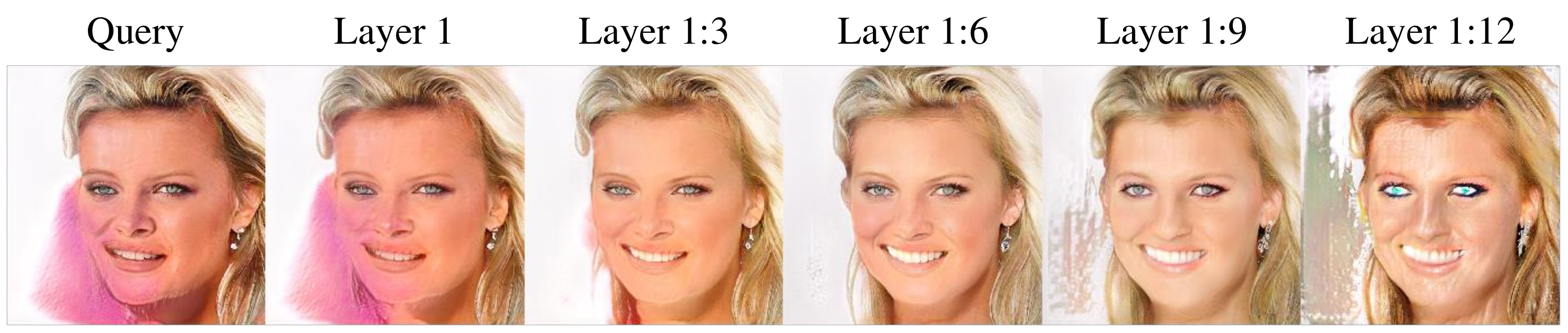}
         \caption{Qualitative comparison for stopping layers}
     \label{fig:rebuttal_celeba_corr_exps}
     \end{subfigure}
     \hfill
     \caption{Quantitative and qualitative results over various hyper parameters (stopping layer $l$ and the portion of ablated units $n$.). We can identify that when the stoppling layer $l$ is set deeper layer, the quality of generation is degraded.}
\end{figure}

 To demonstrate the relation between the hyperparameters and correction performance, we first measure the FID after correction varying the stopping layers $l$ and the portion of ablated units $n$ in PGGAN with CelebA-HQ. In Figure \ref{fig:rebuttal_fid}, ablating 20\% of units in each layer shows the best performance in the FID comparison. The FID sharply increases in layers 9 and 12 whereas the differences are minimal for the lower layers. We also provide another metric called \textit{Realism Score} (R) \cite{Kynkaanniemi2019} which can measure the quality of generation for individual samples for various stopping layers with the fixed $n=20\%$. Figure \ref{fig:rebuttal_realism} indicates the average of $R$ over the 1k artifact generations. We find that the stopping layer 6 shows the best R for the sequential correction.
 
\subsection{Local Region Correction}
\begin{figure}[h!]
    \centering
    \includegraphics[width=0.47\textwidth]{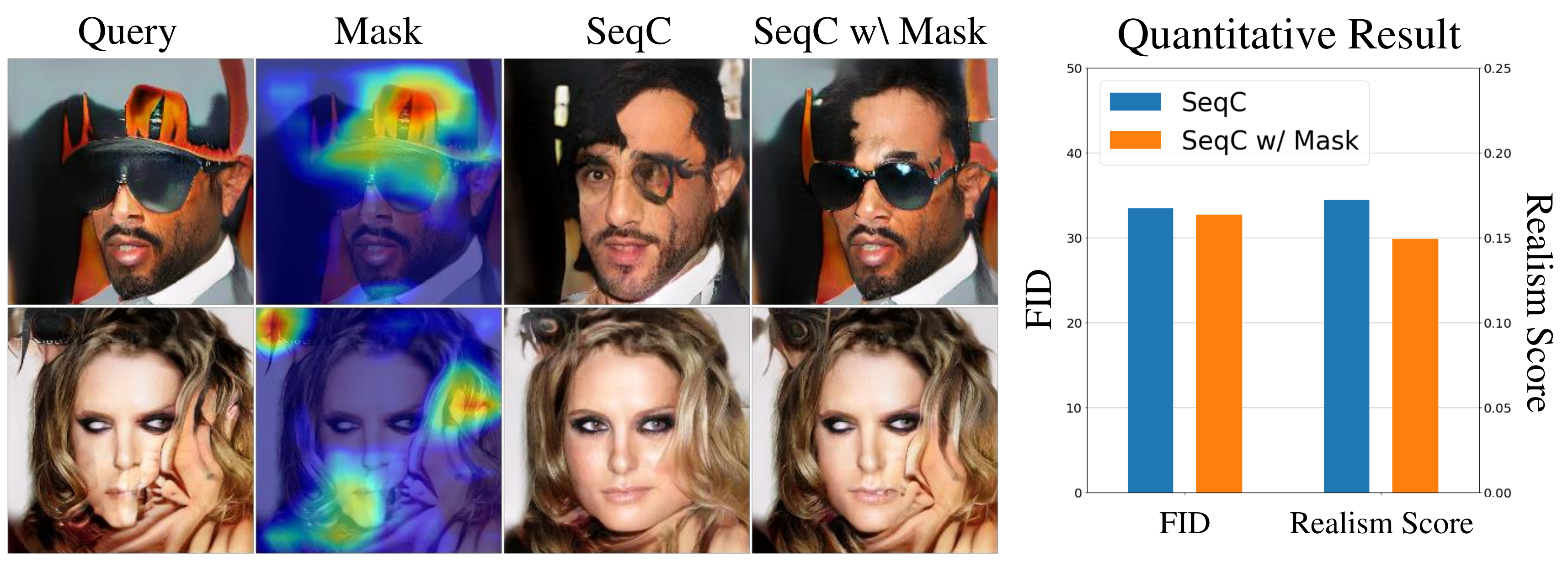}
    \caption{The effect of local region correction for the sequential correction (SeqC) in PGGAN-celebAHQ. We can identify that the area which was not focused on by the mask cannot be changed.}
    \label{fig:local_abl}
\end{figure}

Although the proposed method defines the artifact units in a global sense, we can apply local region correction additionally, since we have the mask of defective regions for individual samples. We change the reducing weight $\lambda (1-DS_{k+1,j,a})$ to $(1-L_a(x))$ where $L_a(x)$ is the downsampled GradCAM mask. In this scheme, we can perform the sample-specified sequential correction and the correction result can be local. As shown in Figure \ref{fig:local_abl}, the local region correction can minimize the change of unmasked contents per individual samples. 

\section{Experimental Evaluations}
This section presents the analytical results of our algorithm and empirical comparisons of various correction methods. We perform correction on three PGGANs trained on LSUN-church, CelebA-HQ, and LSUN-bedroom datasets, respectively. 
We manually label generations from PGGANs and collect 1k artifact generations and its latent codes for each model. 
We obtain the featuremaps through the model and ablate the chosen featuremap units. Throughout the experiments in this paper, we use the stopping layer $l=6$, the number of ablation units $n=20\%$, and the scaling factor $\lambda=0.9$. All the correction experiments are conducted on the same 1k latent codes of the original artifact generations.
The qualitative and quantitative results are presented in the following subsections.

\subsection{Qualitative Result}
\begin{figure*}[t!]
    \centering
        \includegraphics[width=0.96\textwidth]{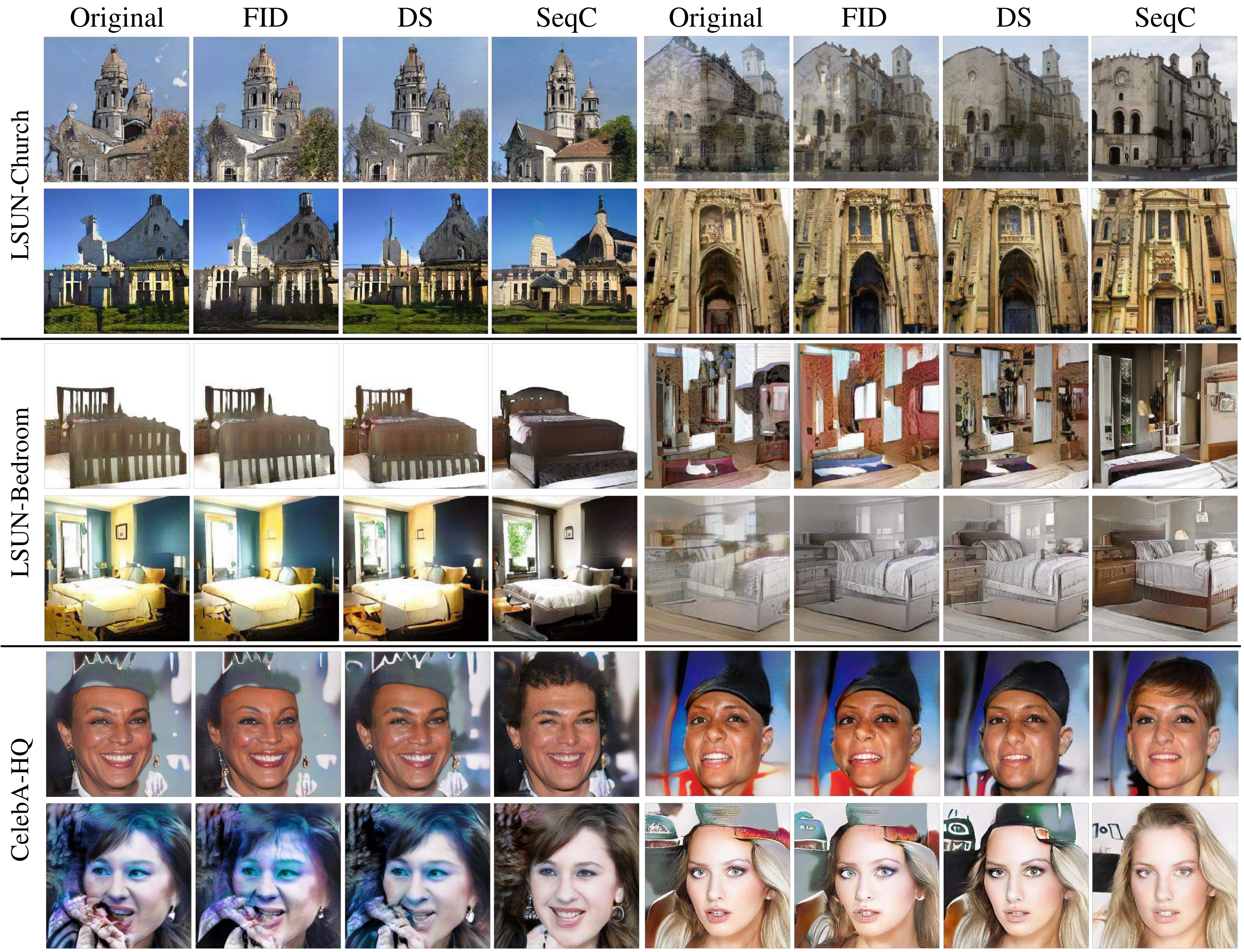}
        \caption{The correction results for each ablation method of the PGGAN with the various dataset (LSUN-Church Outdoor, LSUN-Bedroom, and CelebA-HQ.). We confirm that the sequential correction removes the defection regions effectively.}
        \label{fig:qualitative_result}
\end{figure*}

We first demonstrate how the generations can be corrected for each correction method. As shown in Figure \ref{fig:qualitative_result}, we can identify that the sequential correction can remove the defective region effectively. Especially, ablations in the shallow layer are helpful to correct for the regions in which the generation information is not clear (e.g. the constant green regions in row-5-LHS CelebA-HQ generation) and most information related to the normal characteristics can be maintained. It means that the proposed method minimizes the change of plausible regions and mainly focuses on the defective regions.

\subsection{Quantitative Result}
We use the FID score to quantitatively measure the improvements of artifacts for each approach. In FID calculation, 1k of real training samples are used. 
For better comparison, we report the FID scores on 1k original \textit{artifact} generations and 1k \textit{normal} generations in Table \ref{tab:FID}. As a baseline, we choose random ablation and FID based correction method \cite{bau2018gan} and summarize it under \textit{random} and \textit{FID} row in Table \ref{tab:FID}. Then we ablate the global artifact units obtained by classifier-based approach and summarize it under \textit{DS} row. The results on the sequential correction approach are summarized under \textit{Sequential} row.
The results on each model are summarized under the LSUN-church, CelebA-HQ, and LSUN-bedroom columns, respectively. The proposed method shows the best performance in all three models.

\begin{table}[ht]
\begin{center}
\begin{tabular}{|c|c|c|c|}
    \hline
    \multirow{2}{*}{\textbf{Correction}}&\multirow{2}{*}{\begin{tabular}[c]{@{}c@{}}
    \textbf{LSUN-}\\\textbf{church}\end{tabular}}&\multirow{2}{*}{\begin{tabular}[c]{@{}c@{}}
    \textbf{CelebA-}\\\textbf{HQ}\end{tabular}} &\multirow{2}{*}{\begin{tabular}[c]{@{}c@{}}
    \textbf{LSUN-}\\\textbf{bedroom}\end{tabular}}\\
    &&&\\
    \hline
    \hline
    
Random	&53.43	&42.10	&67.46	\\
\hline
FID	&40.66	&44.37	&48.48	\\
\hline
DS	&32.82	&35.40	&44.93	\\
\hline
Sequential	&\textbf{23.96}	&\textbf{34.71}	&\textbf{40.71}	\\
\hline
\hline
Artifacts &46.95	&36.16	&61.17	\\
\hline
Normals&22.37	&29.80	&29.15	\\
\hline
\end{tabular}
\end{center}
\vskip -0.2in
\caption{FID scores of corrected artifact generations for LSUN-church, CelebA-HQ, and LSUN-bedroom datasets.}
\vskip -0.2in
\label{tab:FID}
\end{table}

\subsection{Human Evaluation}
To support the quality of the corrected results, we provide human evaluation results.
The experiments are constructed by two evaluation procedures: (1) re-labeling the corrected generations from the sequential correction, and (2) assessing the improvement (improved/not improved) for each pre-defined artifact type. We set the criteria for each artifact type to make consistent evaluations. A detailed description for each criterion can be found in Appendix 1.2.

\begin{table}[ht]
\begin{center}
\begin{tabular}{|c|c|c|}
    \hline
    \textbf{Dataset}&\textbf{Corrected (\%)}&\textbf{Improved (\%)}\\
    \hline
    \hline
       CelebA-HQ& 53.00 (4.20)& 96.00 (2.00)\\
       \hline
       LSUN-church& 54.50 (0.90)&  86.10 (6.30)\\
       \hline
       LSUN-bedroom& 46.80 (8.60)& 95.50 (1.10) \\
    \hline
    
\end{tabular}
\end{center}
\vskip -0.2in
\caption{Human evaluation results on the corrected artifact generations. The number in the parentheses indicates the standard deviation over raters.}
\vskip -0.2in
\label{tab:human_evaluation}
\end{table}

The results on 500 artifact corrections for each dataset are summarized in Table \ref{tab:human_evaluation}.
For CelebA-HQ dataset, we could find 53\% out of 500 correction samples obtained by applying our method for artifact-labeled generations are re-labeled as \textit{normal}. While 47\% are still containing artifacts, 97 \% of the total artifact generations have significant improvements in the artifact regions or in the quality of generation. The results on the other two datasets show the similar pattern that almost half of the artifact generations are corrected and the visual quality is mostly improved.

\subsection{Generalization}
Although the proposed method is performed on the generator with a conventional structure, the same approach with minor modification can be generalized for the recent state-of-the-art generators such as StyleGAN2 \cite{Karras2019stylegan2} or U-net GAN \cite{schonfeld2020u} which is a variation of BigGAN \cite{brock2018large}. Because of the distinct structure of each generator (e.g. each generation uses a different convolution kernel in the StyleGAN2), it is non-trivial to align the proposed framework to identify the defective units in the global sense. However, we can obtain the relative defective score for each unit by individually comparing it with the GradCAM mask for implementing the sequential correction. Figure \ref{fig:stylegan_corr_exps} indicates the sequential correction results in the StyleGAN2. More correction examples for StyleGAN2 and U-net GAN are in Appendix 3-4.

\begin{table}[ht]
\begin{center}
\begin{tabular}{|c|c|c|}
    \hline
   \textbf{Model} & \textbf{Artifact} & \textbf{Corrected} \\
    \hline
    \hline
StyleGAN2  	& 117.91	& \textbf{113.08}	\\
\hline
U-net GAN   	&145.15	& \textbf{143.85}	\\
\hline
\end{tabular}
\end{center}
\vskip -0.2in
\caption{FID scores of 100 artifact generations and the corrected generations for StyleGAN2 and U-net GAN on FFHQ.}
\vskip -0.2in
\label{tab:FID_generalization}
\end{table}

\begin{figure}[h!]
    \centering
    \begin{subfigure}[b]{0.94\columnwidth}
    \includegraphics[width=\textwidth]{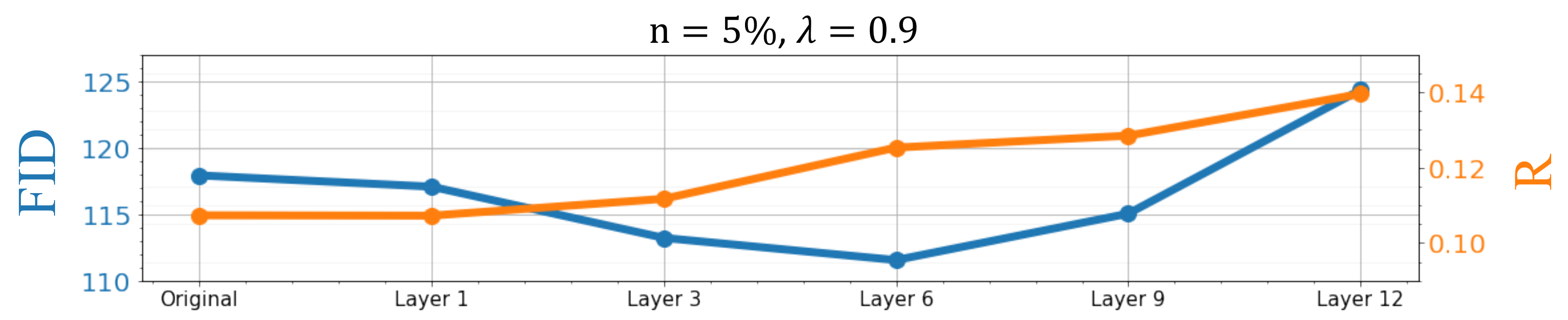}
    \caption{The quantitative results for various stopping layers.}
    \label{fig:stylegan_corr_exps}     
    \end{subfigure}
    
    \begin{subfigure}[b]{0.94\columnwidth}
    \includegraphics[width=\textwidth]{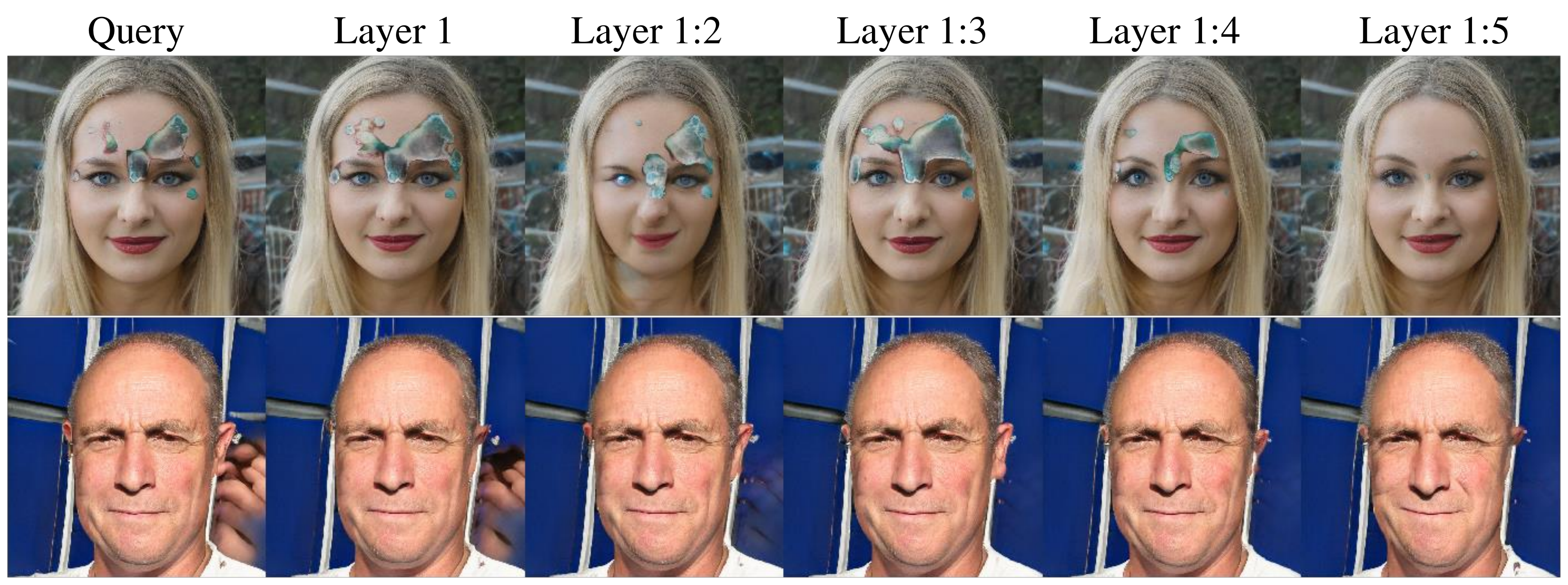}
    \caption{The qualitative results for various stopping layers.}
    \label{fig:stylegan_corr_exps}
    \end{subfigure}
    
    \caption{Correction results on StyleGAN v2 with FFHQ. We select 100 artifact generations and calculate the FID and R for the sequential correction with various stopping layers $l$. We can identify that the scores are improved when the stopping layer $l$ increases up to the middle layer ($l=6$).}
\end{figure}
\section{Discussion}

\begin{figure}[t!]
    \centering
        \includegraphics[width=0.8\columnwidth]{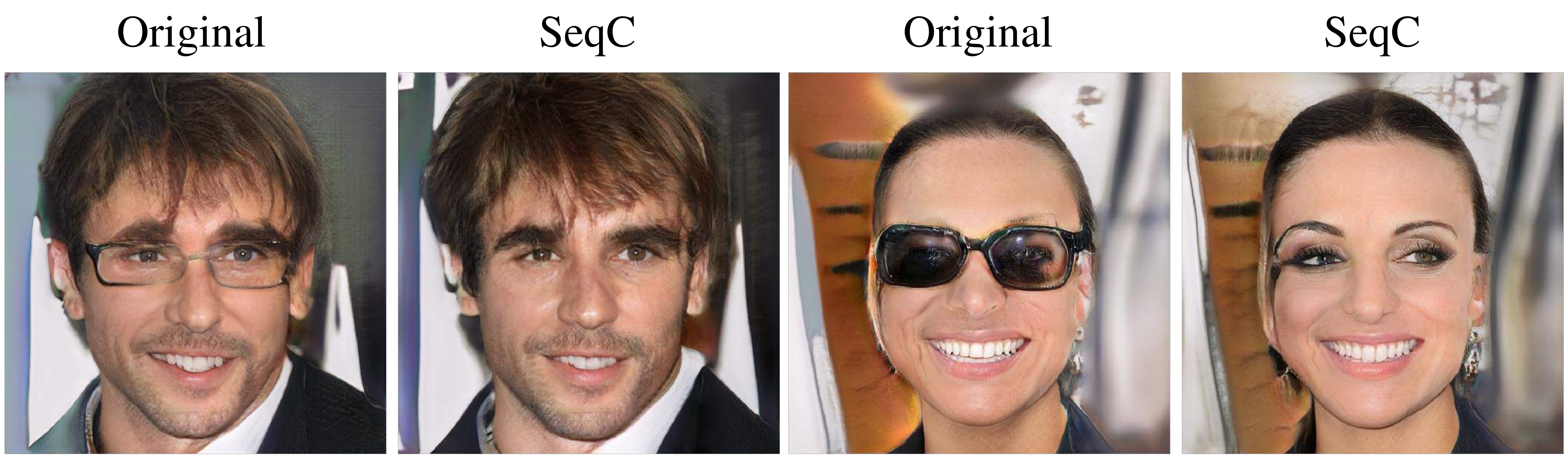}
        \caption{Examples of correction images with glasses or sunglasses. Although the artifact regions are expected background, the sequential correction removes the glasses or sunglasses at the same time.}
        \vskip -0.1in
        \label{fig:sunglass}
\end{figure}

In this paper, we propose the sequential correction method to improve the quality of generation without additional training of the generator. Especially, we define defective score which quantifies relations between each unit and artifact generations with a supervised approach. The sequential correction uses selected units by the DS-based unit identification and adjusts the generation flow in the consequent layers. We show proposed method achieves plausible correction performance and suggest the opportunity of generalization for the various structures of the generator.

While our method has shown to improve the artifact images in terms of both human evaluation and FID-score, there exist some cases that the model improves the image by simplifying the generation. As shown in Figure \ref{fig:sunglass}, we can observe that glasses/sunglasses are removed instead of completing defective regions. We suspect this is caused by the undesirable features that are trained in the classifier which can be further explored in future work.

In addition, while the plausible regions are hardly changing dramatically, we could observe some cases which fail to maintain the original outline of generation. For example, the LSUN-church generation on row-2-RHS in Figure \ref{fig:qualitative_result} shows that the original structure of the church is changed, although the unclear pattern on the door is repaired. In row-6-RHS celebA-HQ generation case, the stains are removed and the blond hair appears, while the angle of the face is changed. We consider this side-effect as a limitation of sequential correction since the identification of artifact units includes average over samples.
For this purpose, the fusion of global and individual sample-based correction can be considered for future work.

\section*{}
\vskip -0.3in
\noindent\textbf{Acknowledgement} 
This work was supported by Institute for IITP grant funded by the Korea government (MSIT) (No.2017-0-01779, XAI and No.2019-0-00075, Artificial Intelligence Graduate School Program (KAIST)).

\newpage
{\small
\bibliographystyle{ieee_fullname}
\bibliography{egbib}
}

\newpage
\onecolumn

\begin{center}
\Large\textbf{Supplementary Material\\Automatic Correction of Internal Units in Generative Neural Networks}
\end{center}
\vskip 1in

\setcounter{section}{0}
\section{Human Evaluations}

\subsection{Evaluation Results}

We select 500 artifacts to measure the performance of corrections based on the criteria in each dataset. The evaluation process is two fold: (1) People re-label for the corrected generation and (2) Check the improvement of generation comparing than original one based on the criteria. The result is summarized in Table \ref{tab:human_evaluation_all}.

\begin{table}[ht]
\begin{center}
\begin{tabular}{|c|c|c|}
    \hline
    \textbf{Dataset}&\textbf{Corrected (\%)}&\textbf{Improved (\%)}\\
    \hline
    \hline
       CelebA-HQ& 53.00 (4.20)& 96.00 (2.00)\\
       \hline
       LSUN-church& 54.50 (0.90)&  86.10 (6.30)\\
       \hline
       LSUN-bedroom& 46.80 (8.60)& 95.50 (1.10) \\
    \hline
    
\end{tabular}
\end{center}
\caption{Human evaluation results on corrected artifact generations. The number in the parentheses indicates the standard deviation over raters.}
\label{tab:human_evaluation_all}
\end{table}

\begin{figure}[h!]
    \centering
    \includegraphics[width=0.94\textwidth]{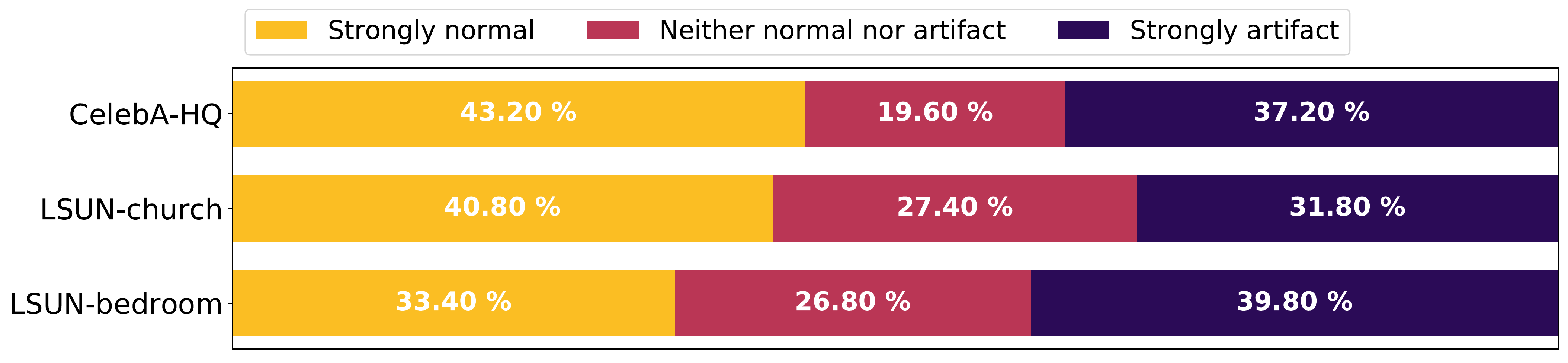}
    \caption{Detailed results of the re-labeling on the corrected images for CelebA-HQ, LSUN-church, and LSUN-bedroom datasets}
    \label{fig:corrected}
\end{figure}

The stacked bar charts in Figure \ref{fig:corrected} shows the human evaluation results on each dataset. `Strongly normal' refers to the portion of the unanimous agreement among 500 samples that the corrected image is normal, and `Strongly artifact' refers to the portion of the unanimous agreement that corrected image is still an artifact. If it is not unanimous agreement, the sample falls into `Neither normal nor aritfact' portion.

\newpage
\subsection{Evaluation Criteria}

\begin{figure}[h!]
    \centering
    \includegraphics[width=0.92\textwidth]{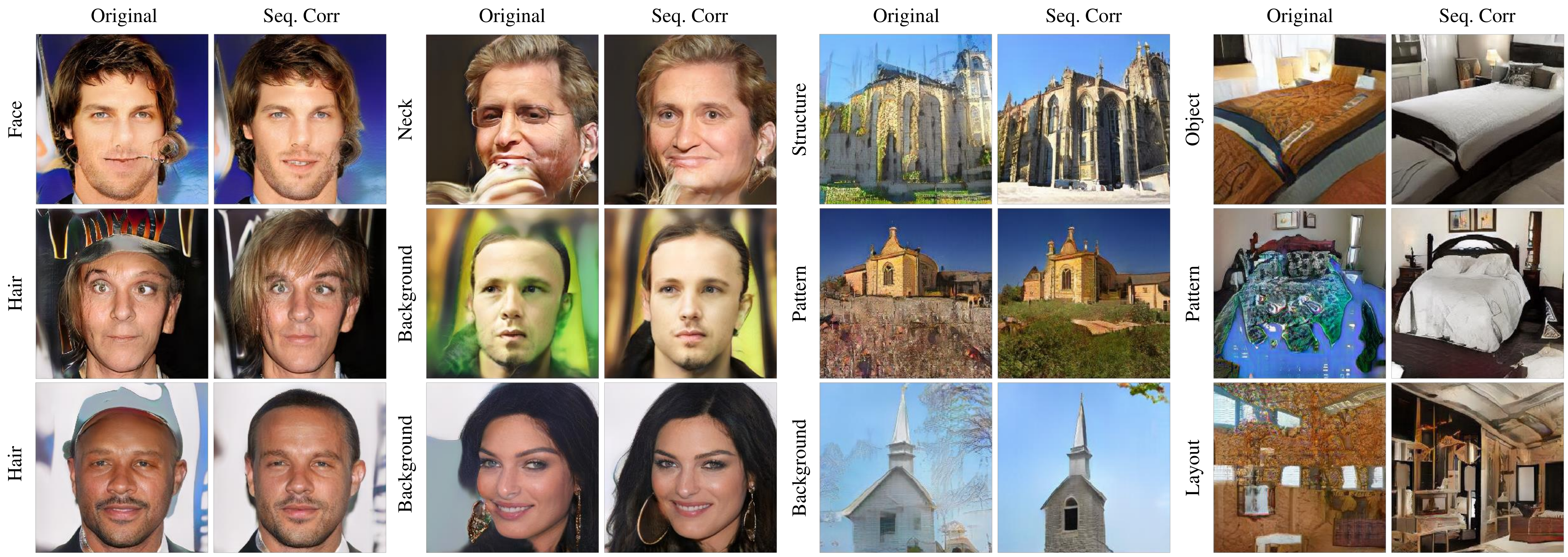}
    \caption{The examples of artifacts based on the descriptions for each dataset. The generation can be included in multiple criteria. For example, left-bottom generation satisfies the hair and background defectiveness at the same time.}
    \label{fig:artifact_label_exp}
\end{figure}

In this section, we describe the criteria for human evaluations. At first, we set the criteria to classify each generation manually. Figure \ref{fig:artifact_label_exp} indicates the examples of artifact depending on each criteria. We set different criteria for each network to handle the main artifact types for each dataset. The detailed descriptions of artifact types and the criteria are summarized in Table \ref{tab:improvements}. 

\begin{table}[ht]
\begin{center}
\begin{tabular}{|c|c|c|}
    \hline
    \textbf{Dataset}&\textbf{Type}&\textbf{Description}\\
    \hline
    \hline
       
   \multirow{3}{*}{CelebA-HQ}   & Face & 
      \begin{minipage} [t] {0.7\columnwidth} 
      
      \begin{itemize}[noitemsep, leftmargin=1em] 
      \item There exist defective segments (i.e., eyes, nose, mouth) on the face. 
      \item The color-tone and texture of the face is not realistic.
     \end{itemize} 
     
    \end{minipage} \\
    
    \cline{2-3}

    & Hair & 
      \begin{minipage} [t] {0.7\columnwidth} 
      \begin{itemize}[noitemsep, leftmargin=1em]

      \item The hair which is fused in the background. 
      \item The head is replaced with patterns. 
      \item There exist empty parts in hair. 
      
     \end{itemize} 
    \end{minipage} \\
    \cline{2-3}
    & Neck & 
      \begin{minipage} [t] {0.7\columnwidth} 
      \begin{itemize}[noitemsep, leftmargin=1em]
      \item The neck is missing. 
      \item Neck and shoulder line is not clear.
      
     \end{itemize} 
    \end{minipage} \\
    
    \cline{2-3}
    & Background & 
      \begin{minipage} [t] {0.7\columnwidth} 
      \begin{itemize}[noitemsep, leftmargin=1em]
      \item There exist clear patterns in the background.

     \end{itemize} 
    \end{minipage} \\
    \hline
    \hline
    
   \multirow{3}{*}{LSUN-church}   & Structural Defects & 
      \begin{minipage} [t] {0.7\columnwidth} 
      
      \begin{itemize}[noitemsep, leftmargin=1em]
      \item Each part of church, i.e., roofs, windows, walls, doors and outlook are not separated each other or from the background.
      \item Church is transparent or overlapped.
      \item The structure of a church is not clear or impossible.
     \end{itemize} 
     
    \end{minipage} \\
    
    \cline{2-3}

    & Pattern & 
      \begin{minipage} [t] {0.7\columnwidth} 
      \begin{itemize}[noitemsep, leftmargin=1em]

      \item There exists a clear pattern (i.e., checkerboard pattern) which is unnatural. 
      \item The color of an object is unnatural.

     \end{itemize} 
    \end{minipage} \\
    \cline{2-3}
    & Background & 
      \begin{minipage} [t] {0.7\columnwidth} 
      \begin{itemize}[noitemsep, leftmargin=1em]
      \item There exists standing out, unrecognizable object. 
      \item There are transparent objects in the sky.
     \end{itemize} 
    \end{minipage} \\
    
    \hline
    \hline

   \multirow{3}{*}{LSUN-bedroom}   & Object & 
      \begin{minipage} [t] {0.7\columnwidth} 
      
      \begin{itemize}[noitemsep, leftmargin=1em]
      \item There is no bed in bedroom. 
      \item The structure or edge of an object is physically not feasible.
     \end{itemize} 
     
    \end{minipage} \\
    
    \cline{2-3}

    & Pattern & 
      \begin{minipage} [t] {0.7\columnwidth} 
      \begin{itemize}[noitemsep, leftmargin=1em]

      \item The texture of the images is like a painting. 
      \item There is a transparent patterns.  
      \item There is a clear pattern crossing the boundary of an object.

     \end{itemize} 
    \end{minipage} \\
    \cline{2-3}
    & Layout & 
      \begin{minipage} [t] {0.7\columnwidth} 
      \begin{itemize}[noitemsep, leftmargin=1em]
      \item The layout of the bedroom is not clear and realistic (i.e., The walls are not connected).
      \item No objects or room at all. 
      
     \end{itemize} 
    \end{minipage} \\
    \hline

\end{tabular}
\end{center}
\caption{Types of artifact generations and its description}
\label{tab:improvements}
\end{table}

\newpage

\section{Automatic Correction Results}
\begin{figure}[h!]
    \centering
    \includegraphics[width=0.9\textwidth]{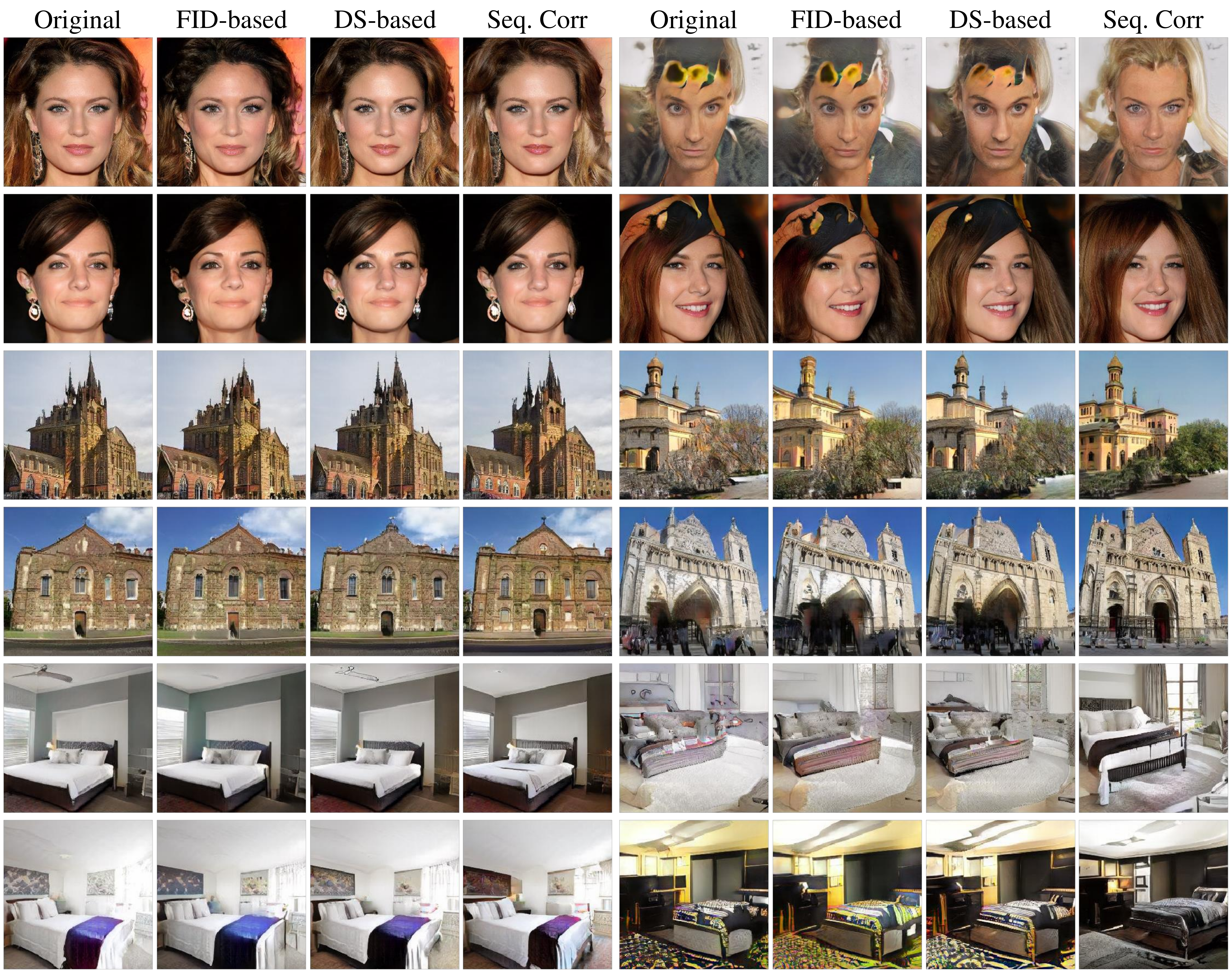}
    \caption{Correction result for each method in various dataset. Although the entire method can conserve the plausible regions, the proposed method can correct the defective regions effectively.}
    \label{fig:summary}
\end{figure}

\begin{figure}[h!]
    \centering
    \includegraphics[width=0.9\textwidth]{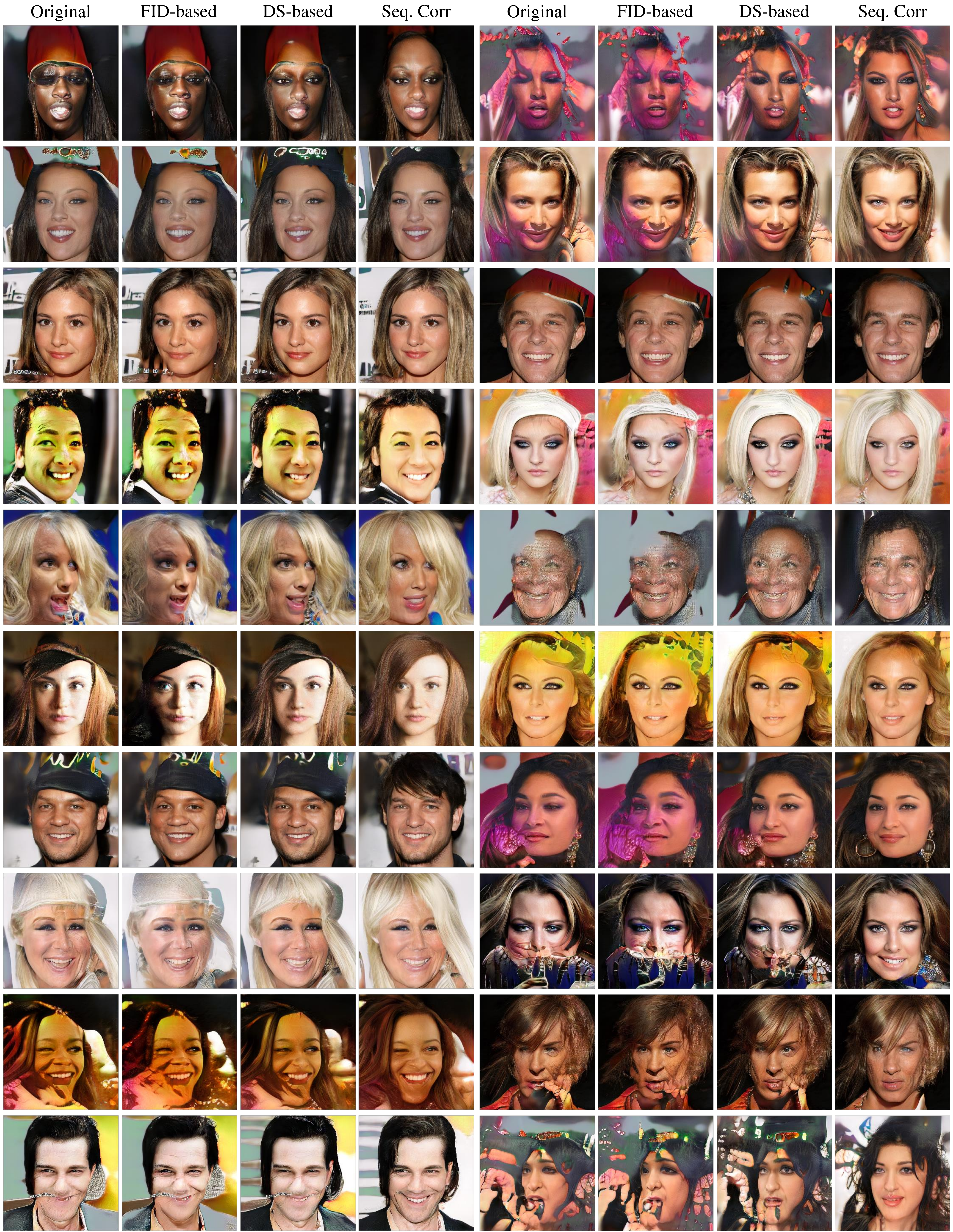}
    \caption{Correction results for the artifacts in PGGAN with CelebA-HQ.}
    \label{fig:celeba_corr}
\end{figure}

\begin{figure}[h!]
    \centering
    \includegraphics[width=0.9\textwidth]{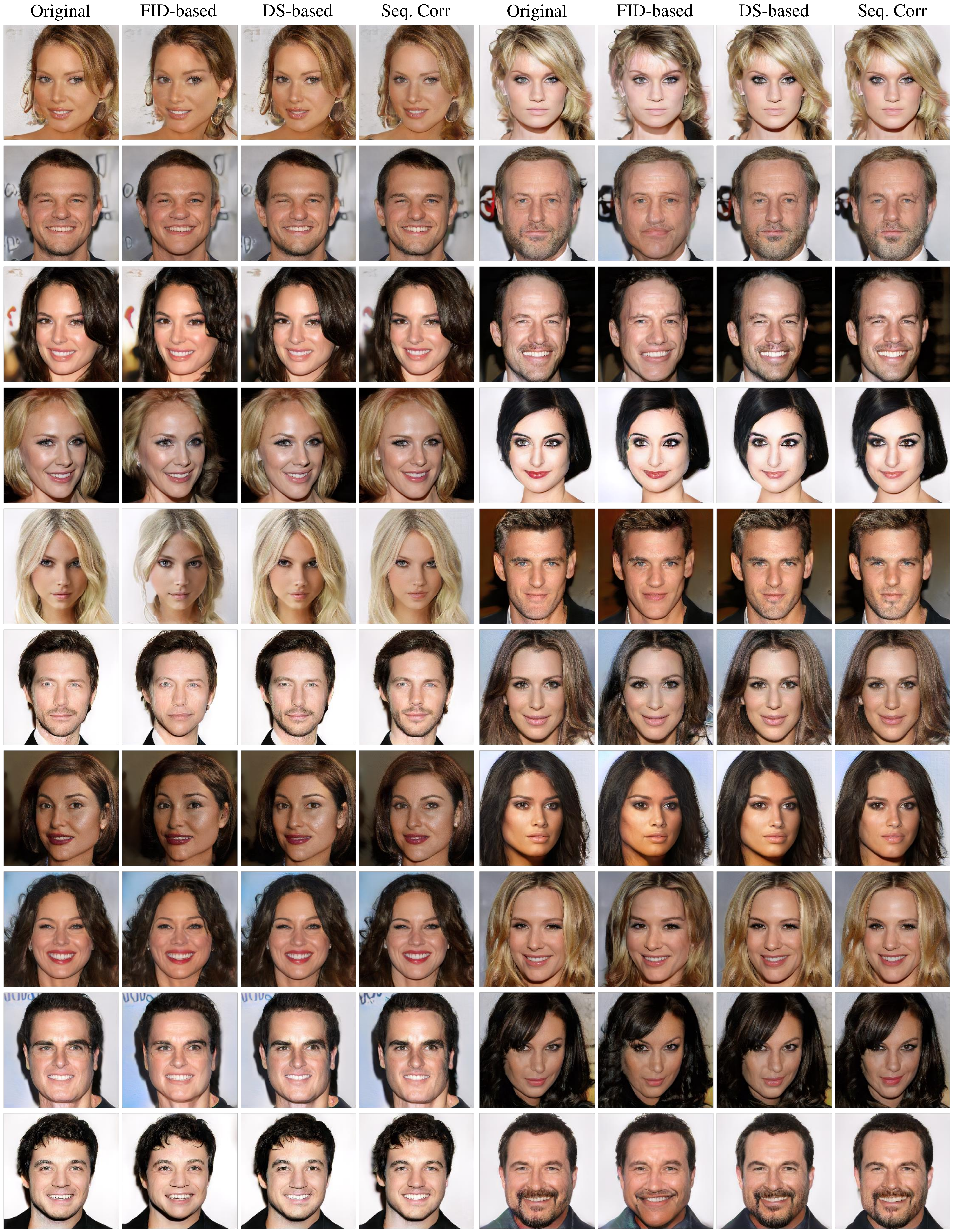}
    \caption{Correction results for the normal generations in PGGAN with CelebA-HQ.}
    \label{fig:celeba_corr_normal}
\end{figure}

\begin{figure}[h!]
    \centering
    \includegraphics[width=0.9\textwidth]{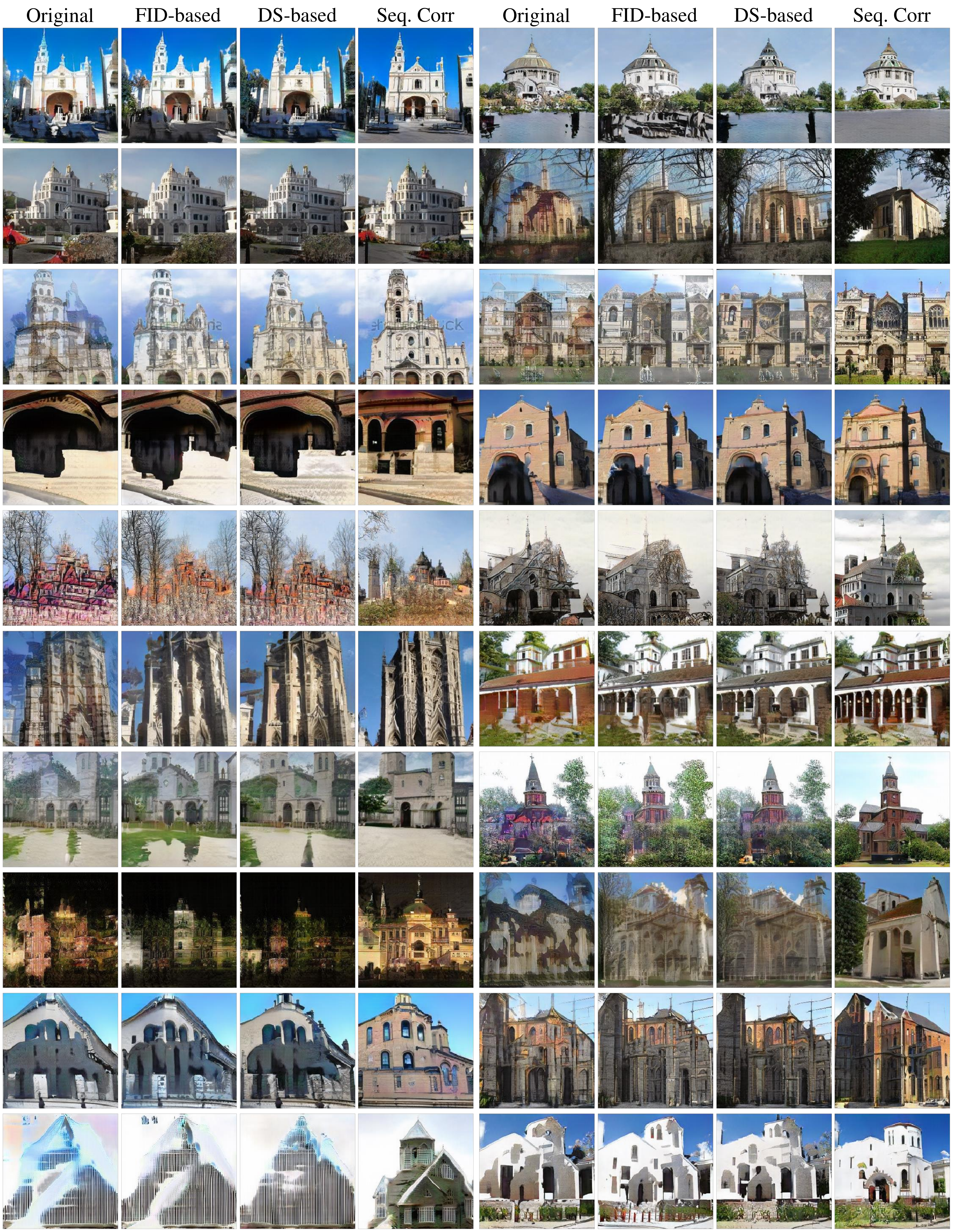}
    \caption{Correction results for the artifacts in PGGAN with LSUN-church Outdoor.}
    \label{fig:church_corr}
\end{figure}

\begin{figure}[h!]
    \centering
    \includegraphics[width=0.9\textwidth]{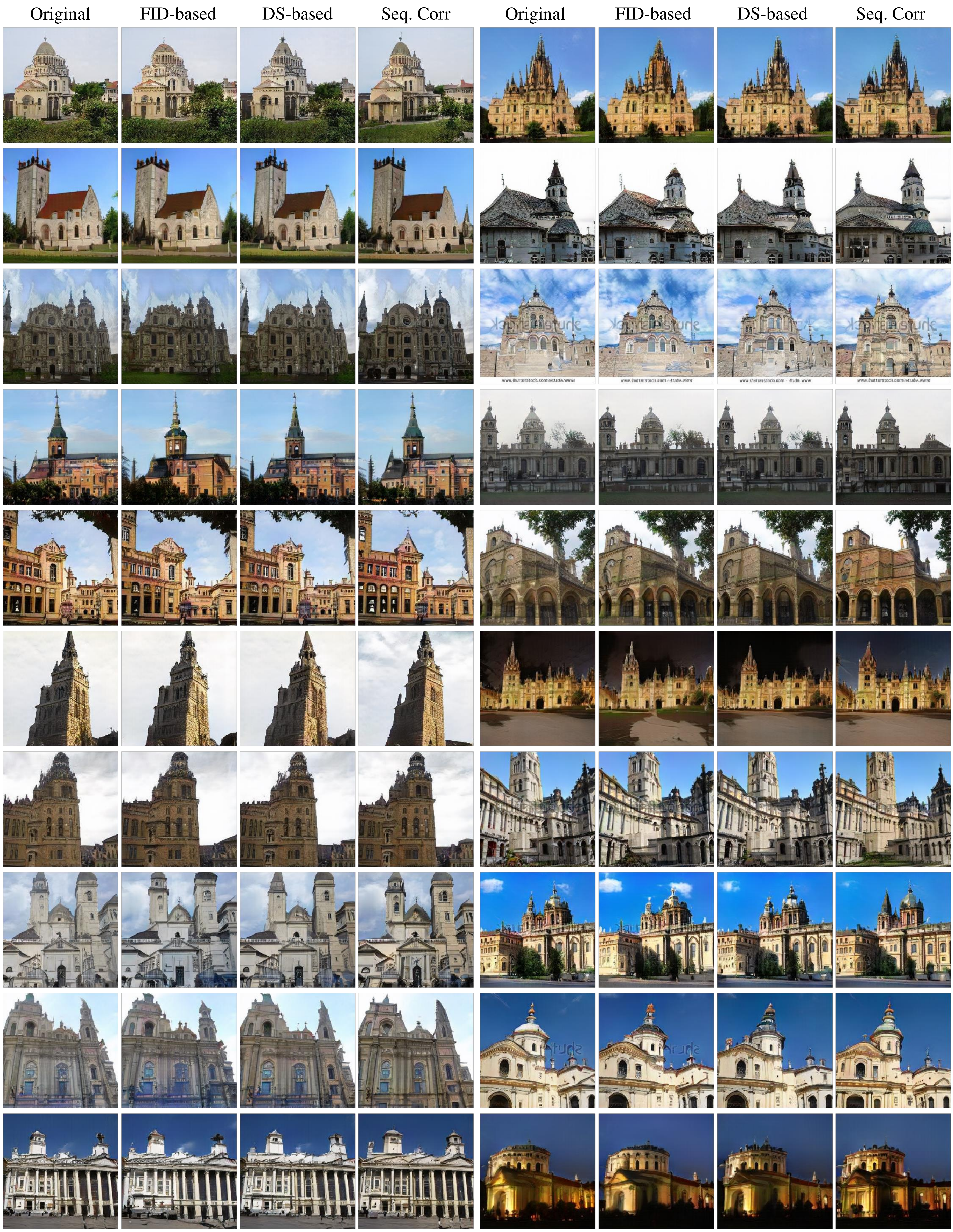}
    \caption{Correction results for the normal generations in PGGAN with LSUN-church Outdoor.}
    \label{fig:church_corr_normal}
\end{figure}

\begin{figure}[h!]
    \centering
    \includegraphics[width=0.9\textwidth]{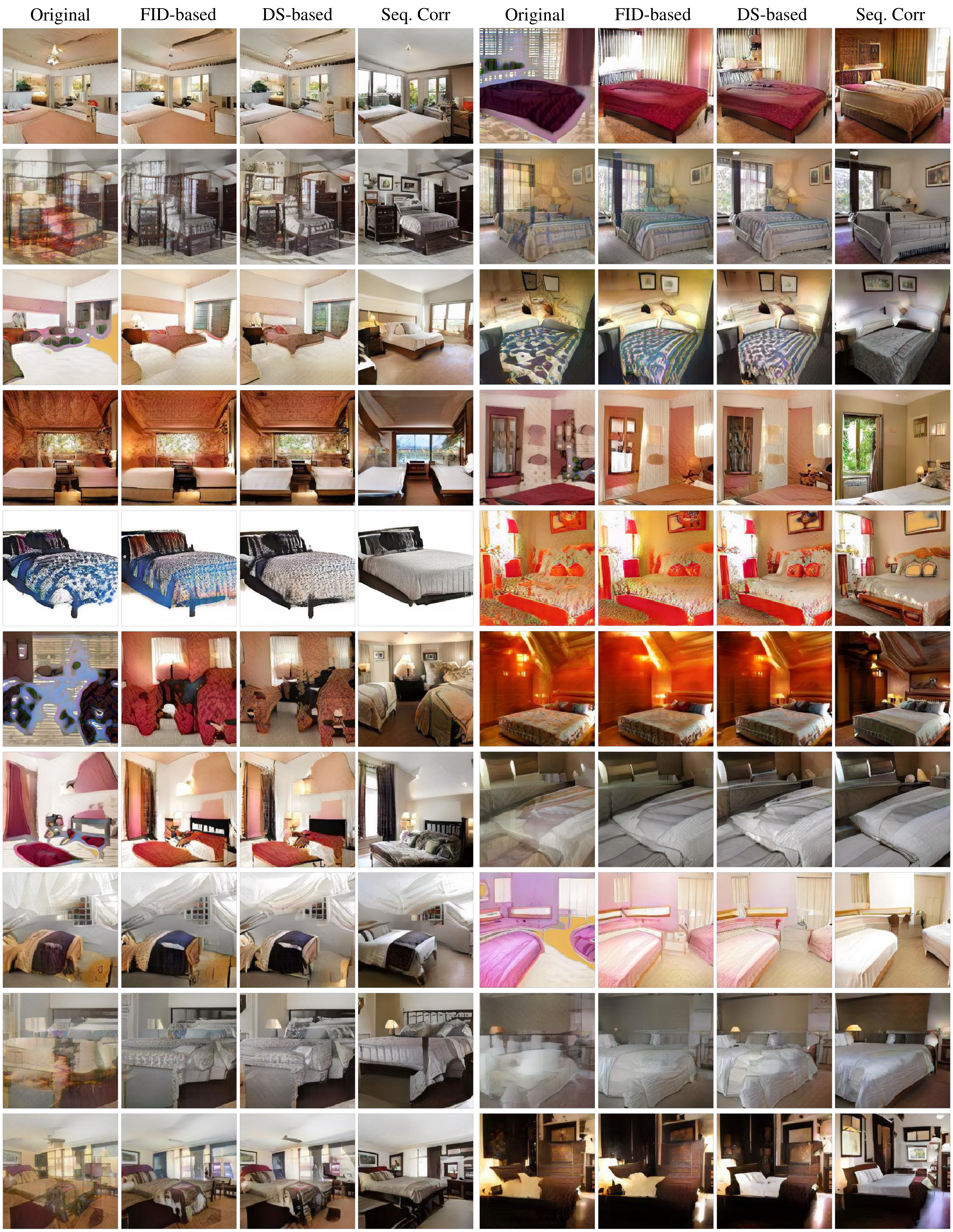}
    \caption{Correction results for the artifacts in PGGAN with LSUN-bedroom.}
    \label{fig:bedroom_corr}
\end{figure}

\begin{figure}[h!]
    \centering
    \includegraphics[width=0.9\textwidth]{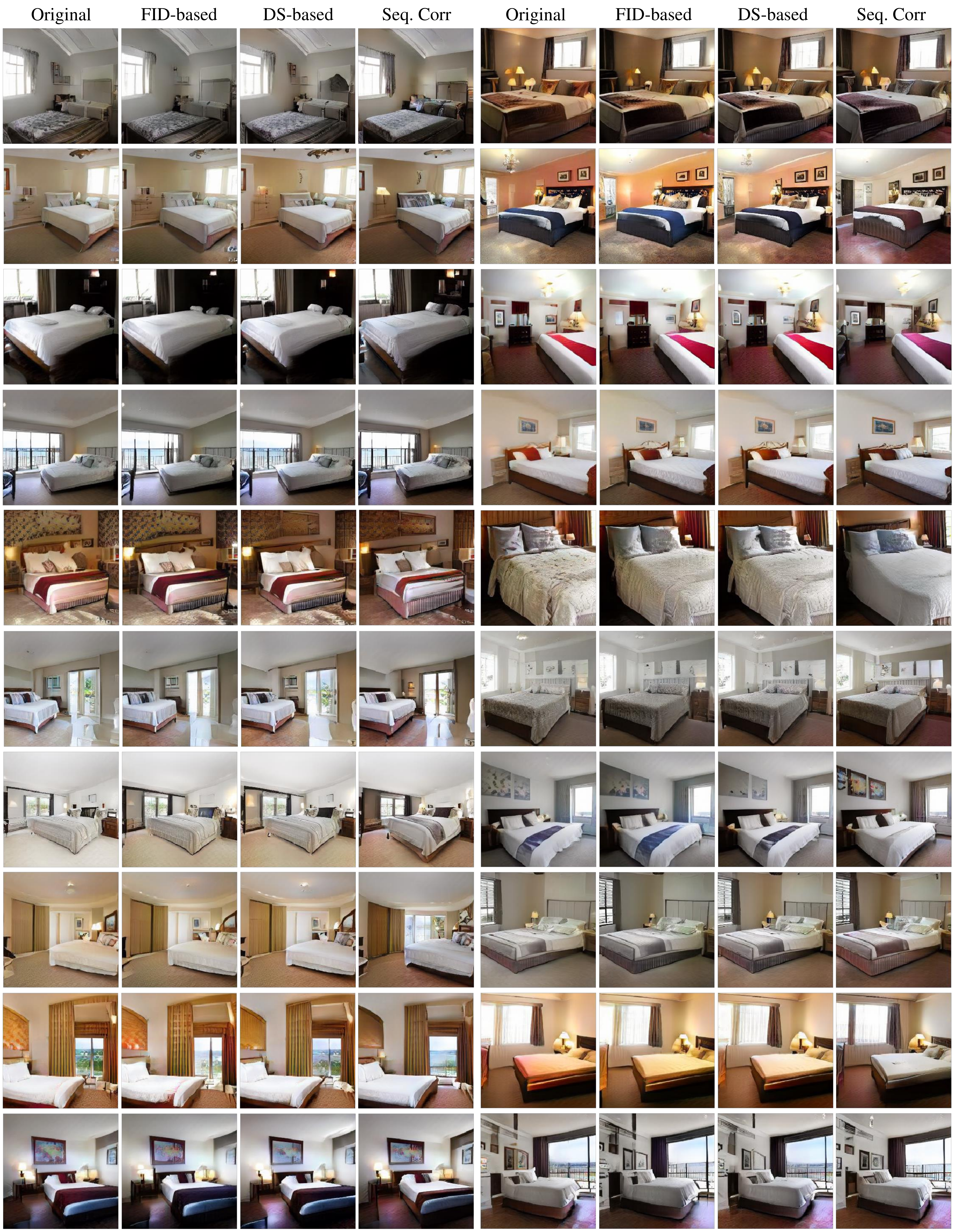}
    \caption{Correction results for the normal generations in PGGAN with LSUN-bedroom.}
    \label{fig:bedroom_corr_normal}
\end{figure}

We visualize the correction results for each method. FID and defective scores (DS) based correction methods are applied in layer 6. The sequential correction is applied from layer 1 to 6 (The layer 0 means the first dense layer with reshape.). Figure \ref{fig:summary} indicates the correction result for random generations. Although we can identify that all of methods maintain information for the plausible regions, the sequential correction method shows effective repair for the defective regions (e.g. The stain on the forehead can be removed from the sequential correction method (Top-right in Figure \ref{fig:summary}.). Note that all of correction methods are in the global approach to improve the generator not the repair per query (given generation). The entire methods basically stop or reduce the generation information in the network (The indices of units are pre-defined based on FID score or DS.). Figure \ref{fig:celeba_corr} - \ref{fig:bedroom_corr_normal} indicate the correction results for each class in various dataset.

\newpage
\subsection{Artifacts in PGGAN with CelebA-HQ}

\newpage
\subsection{Normal in PGGAN with CelebA-HQ}

\newpage
\subsection{Artifacts in PGGAN with LSUN-church Outdoor}

\subsection{Normal in PGGAN with LSUN-church Outdoor}

\subsection{Artifacts in PGGAN with LSUN-bedroom}

\subsection{Normal in PGGAN with LSUN-bedroom}

\section{Sequential Correction in StyleGAN v2}
\begin{figure}[h!]
    \centering
    \includegraphics[width=0.7\textwidth]{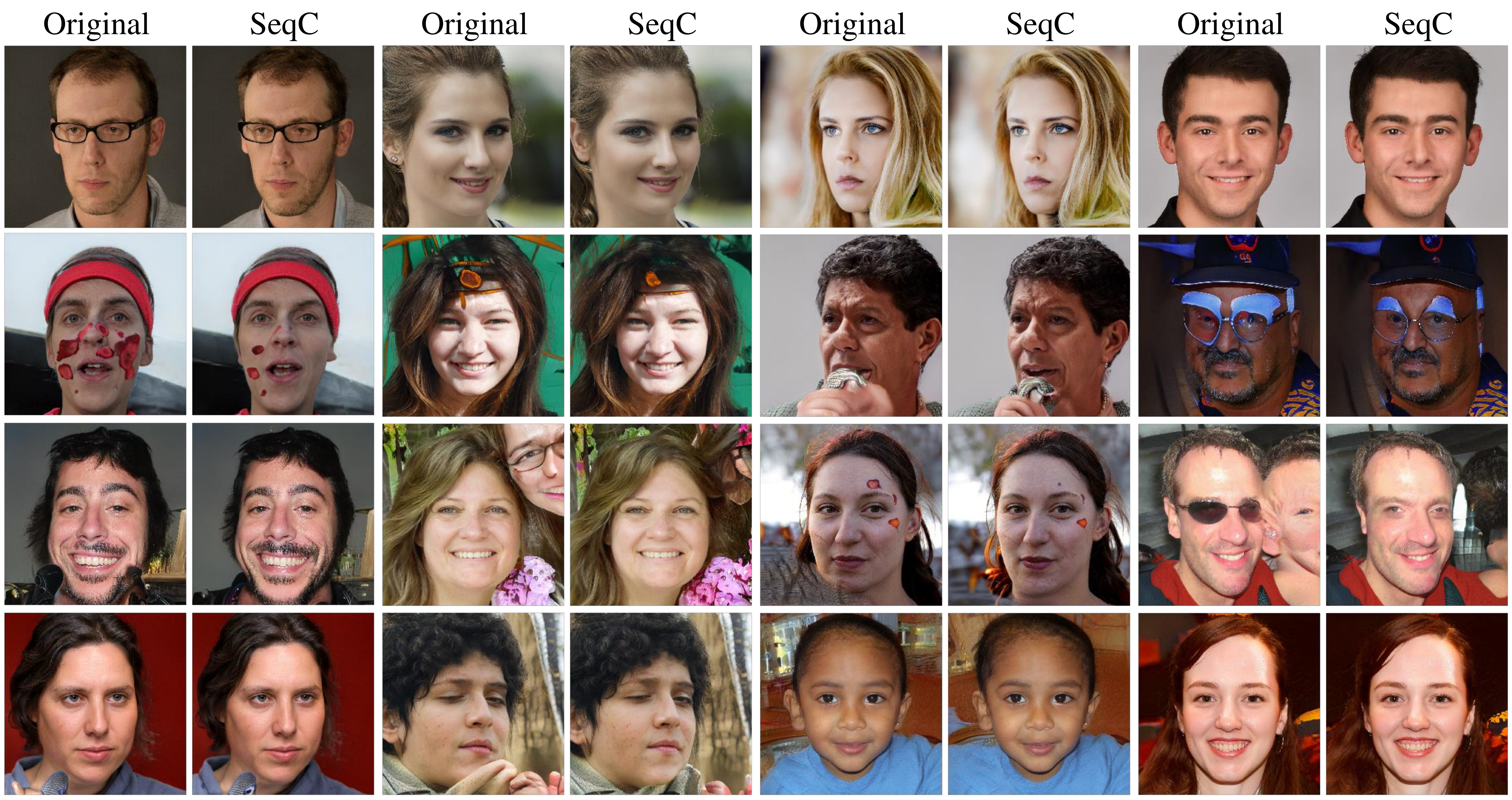}
    \caption{Correction result for the generations in StyleGAN v2 with FFHQ.}
    \label{fig:styleganv2_seqc}
\end{figure}

Due to the structural difference between PGGAN and StyleGAN v2(e.g. stochastic variation induced at each convolutional layer), the global unit correction is not trivial. However, we modified our method to use each generation artifact map obtained by GradCAM to derive the sequential correction. After a generation is classified as artifact, we obtain a mask for defective region. Note that the classifier is only trained once using the PGGAN generated annotated images on CelebAHQ. Similar to our global unit identification for PGGAN, we can assign relative defective score using GradCAM mask. Figure \ref{fig:styleganv2_seqc} depicts the result of sequential local region correction for images generated by StyleGAN v2 trained on FFHQ dataset.

\section{Sequential Correction in U-net GAN}
As described in Section 5.3 in the paper, we can apply our method with some modification to other sate-of-the-art generation models. For this purpose, we select U-net GAN which the generator module is same with BigGAN with some modifications. Specifically, U-Net GAN generator feed the same input latent vector to BatchNorm. Furthermore, they introduced an unconditional generator by replacing class-conditional BatchNorm in BigGAN with self-modulation. We followed the same procedure as StyleGAN v2 for sequential correction.
\begin{figure}[h!]
    \centering
    \includegraphics[width=0.7\textwidth]{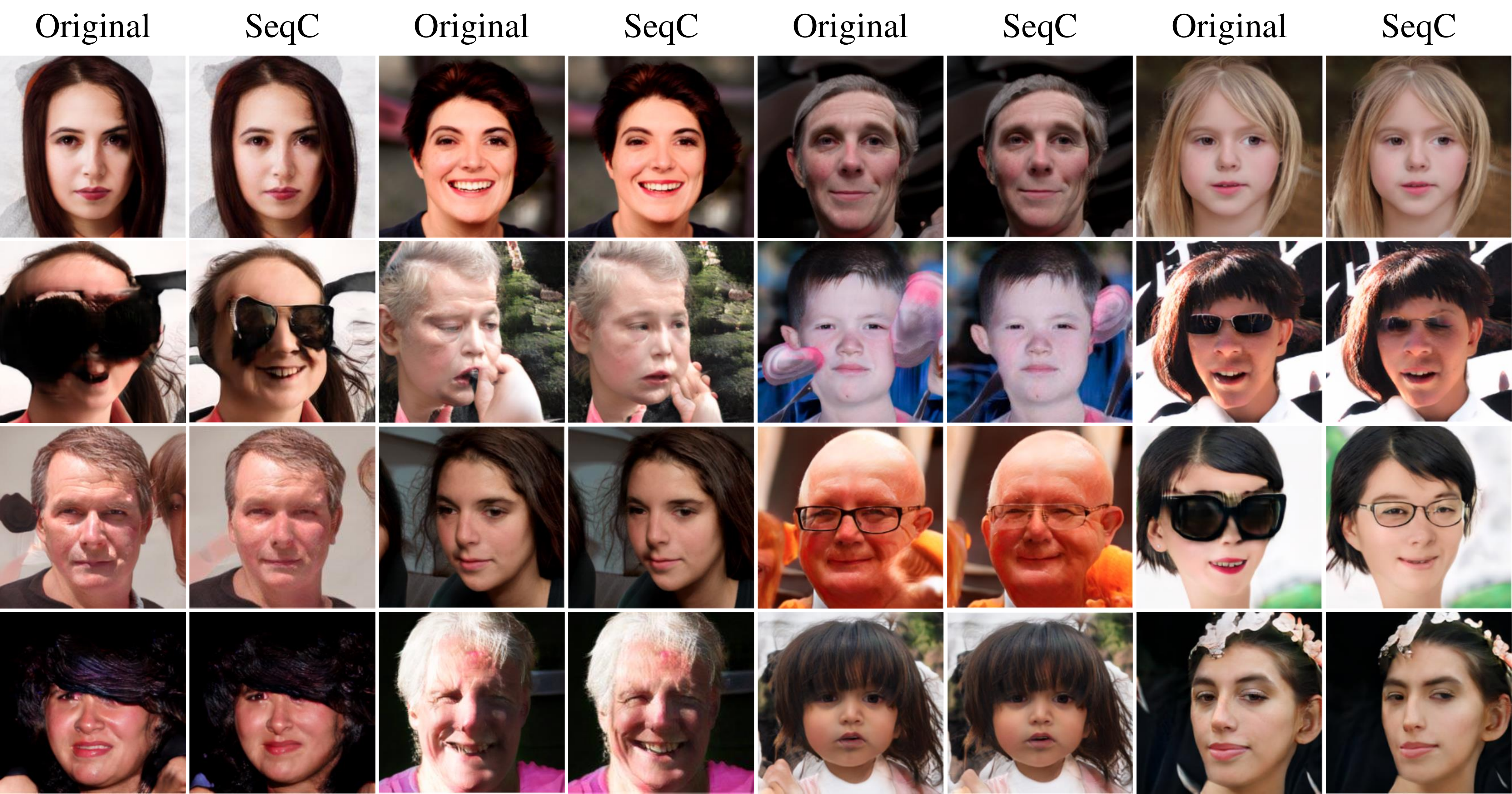}
    \caption{Correction result for the generations in U-net GAN with FFHQ.}
    \label{fig:unetgan_seqc}
\end{figure}

\newpage
\section{Representative generation for each unit}
\begin{figure}[h!]
    \centering
    \includegraphics[width=0.9\textwidth]{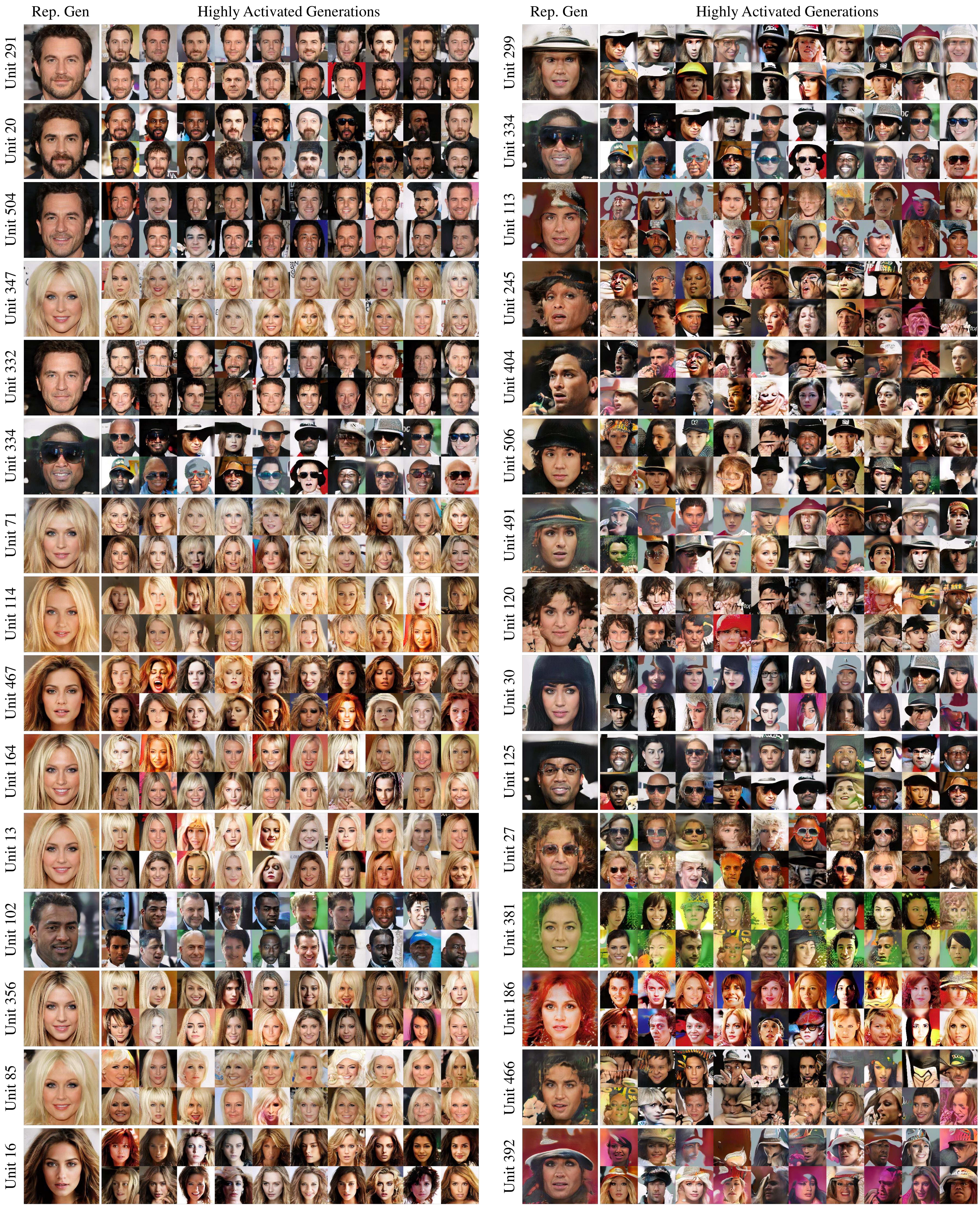}
    \caption{Representative generation for each unit in layer 6 of PGGAN with CelebA-HQ. (Left) FID score based unit selections. (Right) DS based unit selections.}
    \label{fig:unit_gen_celeba}
\end{figure}

\begin{figure}[h!]
    \centering
    \includegraphics[width=0.9\textwidth]{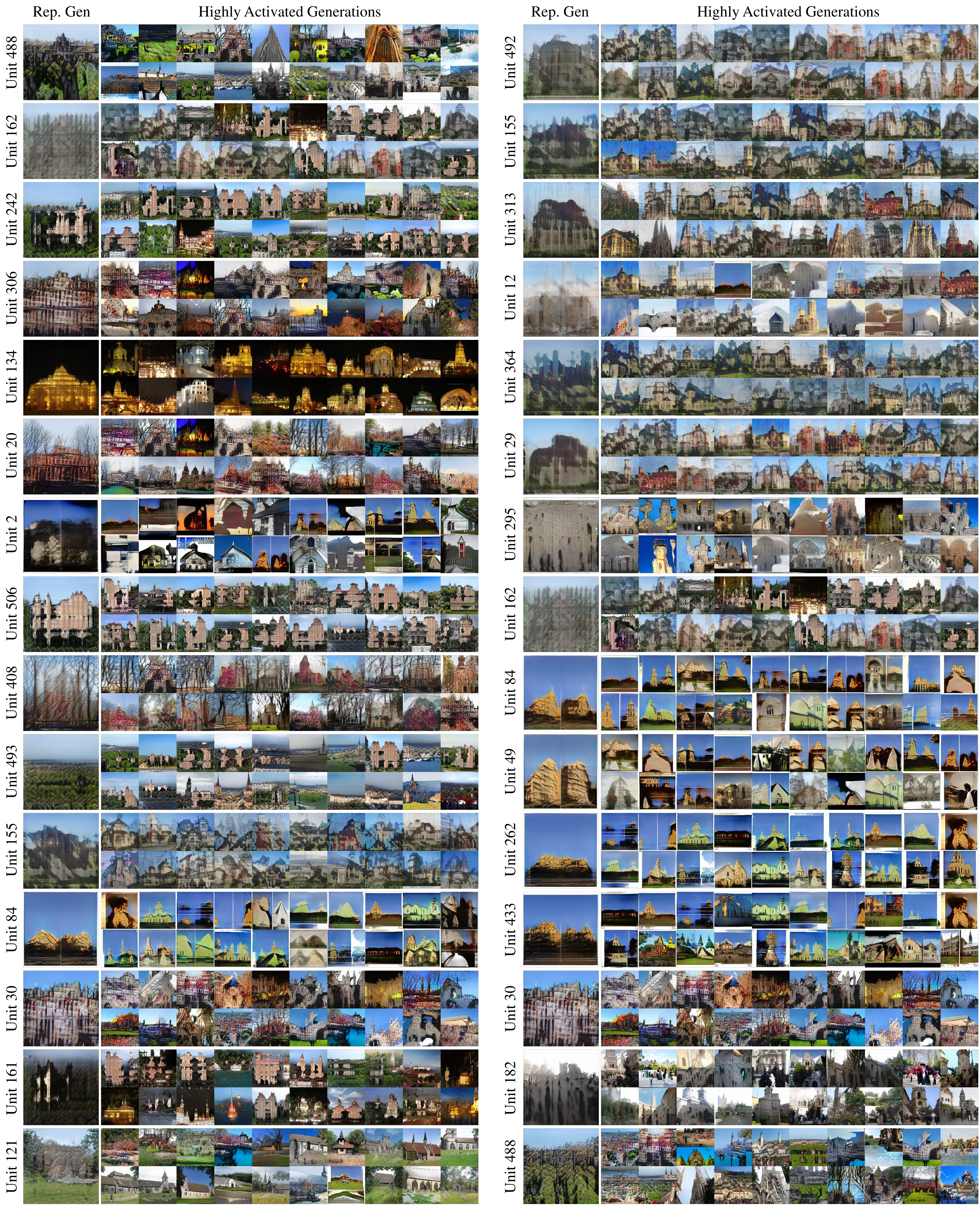}
    \caption{Representative generation for each unit in layer 6 of PGGAN with LSUN-church Outdoor. (Left) FID score based unit selections. (Right) DS based unit selections.}
    \label{fig:unit_gen_church}
\end{figure}

\begin{figure}[h!]
    \centering
    \includegraphics[width=0.9\textwidth]{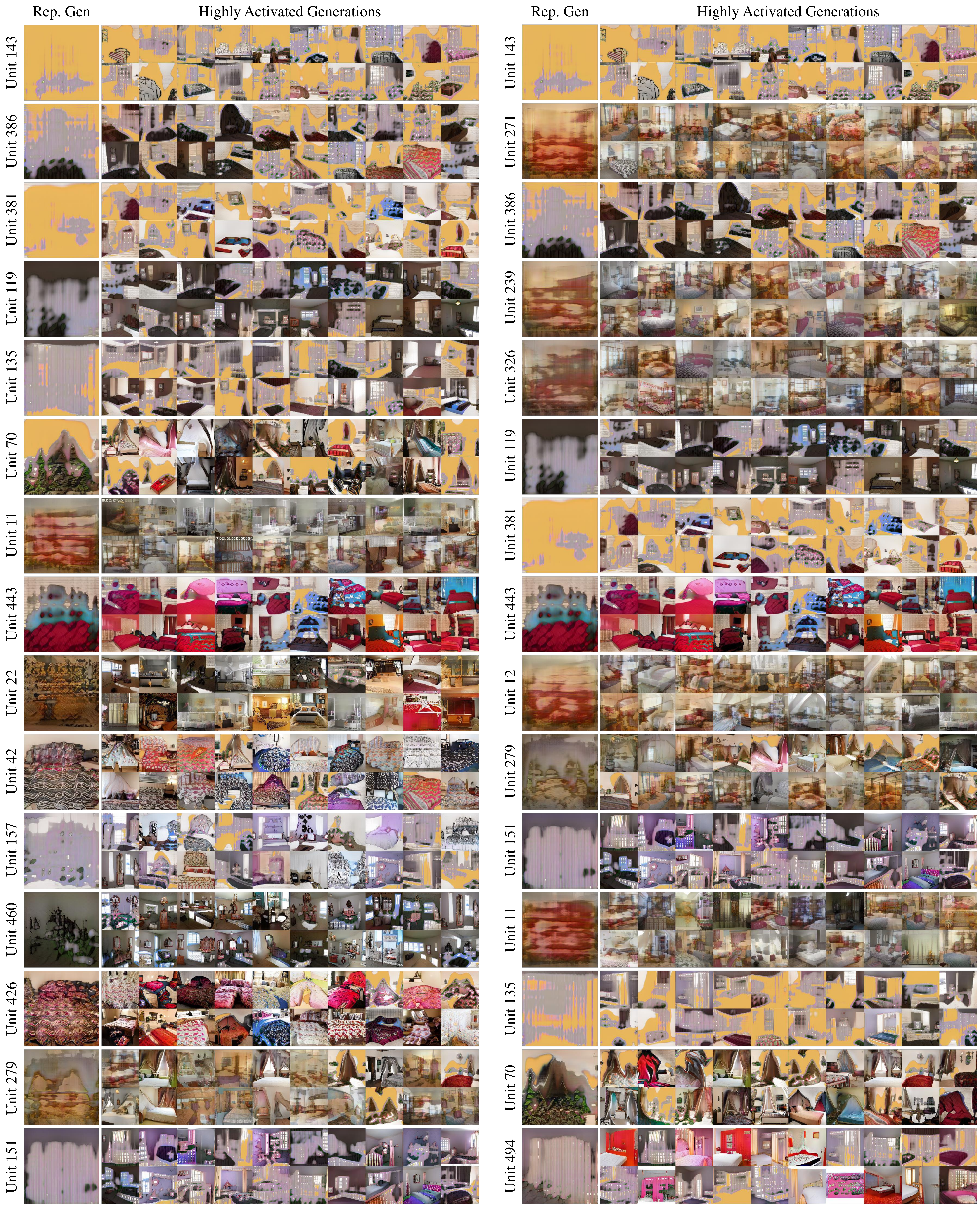}
    \caption{Representative generation for each unit in layer 6 of PGGAN with LSUN-bedroom. (Left) FID score based unit selections. (Right) DS based unit selections.}
    \label{fig:unit_gen_bedroom}
\end{figure}

In this section, we visualize the the representative generation and highly activated generations for selected units in layer 6 of PGGAN generator with various dataset to validate the selected units generation concepts. Figure \ref{fig:unit_gen_celeba} - \ref{fig:unit_gen_bedroom} indicate the visualization results from top 15 units for each scoring method (left: FID scores and right: defective score (DS)). 

\begin{figure}[h!]
    \centering
    \includegraphics[width=0.9\textwidth]{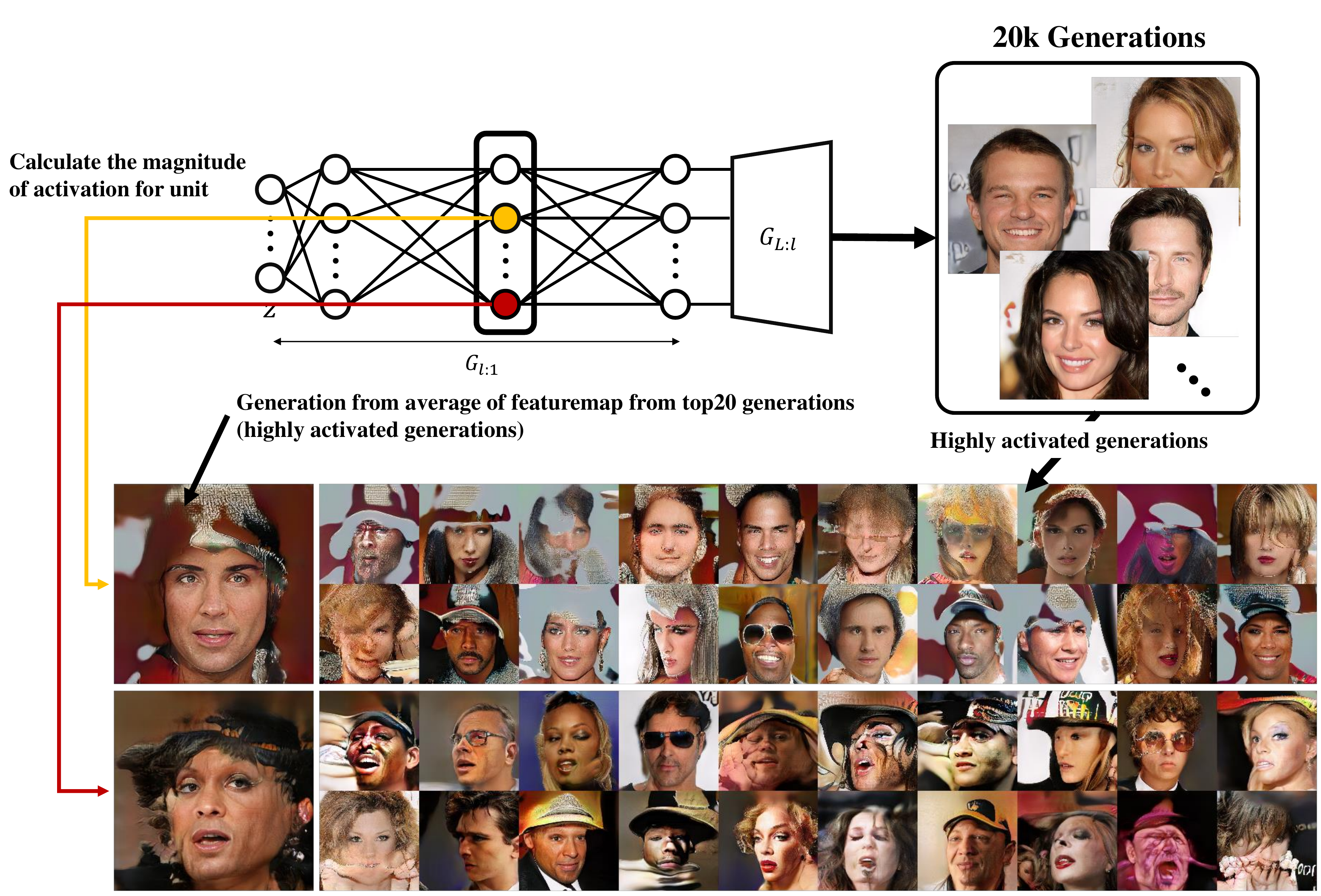}
    \caption{How to select 20 generations that maximize each featuremap unit }
    \label{fig:unit_selection}
\end{figure}

\end{document}